\documentclass[10pt,twocolumn,letterpaper]{article}

\usepackage[pagenumbers]{cvpr} % To force page numbers, e.g. for an arXiv version

\usepackage[table,dvipsnames]{xcolor}

\usepackage{xspace}

\newcommand{\alias}{SkillDiffuser\xspace}
\newcommand{\lorl}{LOReL\xspace}
\newcommand{\metaworld}{Meta-World\xspace}

\definecolor{bestcolor}{gray}{.9}
\newcommand{\bestcell}[1]{\cellcolor{bestcolor}{#1}}

\newcommand{\bs}{\boldsymbol{s}}
\newcommand{\ba}{\boldsymbol{a}}

\newcommand{\btau}{\boldsymbol{\tau}}
\newcommand{\by}{\boldsymbol{y}}

\usepackage{changepage}

\newcommand{\argmin}{\operatornamewithlimits{arg\,min}}
\newcommand{\vct}[1]{\boldsymbol{#1}}

\usepackage{amsmath}
\usepackage{amssymb}
\usepackage{mathtools}
\usepackage{pifont}
\usepackage{amsthm}
\theoremstyle{plain}
\newtheorem{theorem}{Theorem}[section]

\theoremstyle{definition}

\theoremstyle{remark}

\usepackage{algorithm}
\usepackage{algpseudocode}
\algrenewcommand\algorithmicrequire{\textbf{Input:}}
\algrenewcommand\algorithmicensure{\textbf{Output:}}

\definecolor{cvprblue}{rgb}{0.21,0.49,0.74}
\usepackage[pagebackref,breaklinks,colorlinks,citecolor=cvprblue]{hyperref}
\usepackage[accsupp]{axessibility} % Improves PDF readability for those with visual impairments.

\def\paperID{9305} % *** Enter the Paper ID here
\def\confName{CVPR}
\def\confYear{2024}

\title{SkillDiffuser: Interpretable Hierarchical Planning via Skill Abstractions in Diffusion-Based Task Execution}

\author{
Zhixuan Liang$^{1,3}$ \quad
Yao Mu$^{1,3}$ \quad 
Hengbo Ma$^2$ \quad \\
Masayoshi Tomizuka$^2$ \quad 
Mingyu Ding$^{2}$\footnotemark[2] \quad 
Ping Luo$^{1,3}$\footnotemark[2]\\
[1.5mm]
$^1$The University of Hong Kong \quad
$^2$University of California, Berkeley \quad
$^3$Shanghai AI Laboratory \quad\\
{\tt\small \{zxliang, ymu, pluo\}@cs.hku.hk \quad \{hengbo\_ma, tomizuka, myding\}@berkeley.edu}
\\
\normalsize{\url{https://skilldiffuser.github.io/}}
}
\begin{document}
\maketitle
% \footnotetext[1]{Equal contribution.}
\footnotetext[2]{Corresponding authors.}

\begin{abstract}
Diffusion models have demonstrated strong potential for robotic trajectory planning. However, generating coherent trajectories from high-level instructions remains challenging, especially for long-range composition tasks requiring multiple sequential skills. 
We propose \alias, an end-to-end hierarchical planning framework integrating interpretable skill learning with conditional diffusion planning to address this problem. 
At the higher level, the skill abstraction module learns discrete, human-understandable skill representations from visual observations and language instructions. These learned skill embeddings are then used to condition the diffusion model to generate customized latent trajectories aligned with the skills. This allows generating diverse state trajectories that adhere to the learnable skills.
By integrating skill learning with conditional trajectory generation, SkillDiffuser produces coherent behavior following abstract instructions across diverse tasks.
Experiments on multi-task robotic manipulation benchmarks like \metaworld and \lorl demonstrate state-of-the-art performance and human-interpretable skill representations from \alias. 
More visualization results and information could be found on our \href{https://skilldiffuser.github.io/}{website}.

\end{abstract}    
\section{Introduction}
\label{sec:intro}

\begin{figure}[tb]
\centering 
\includegraphics[width=0.99\linewidth]{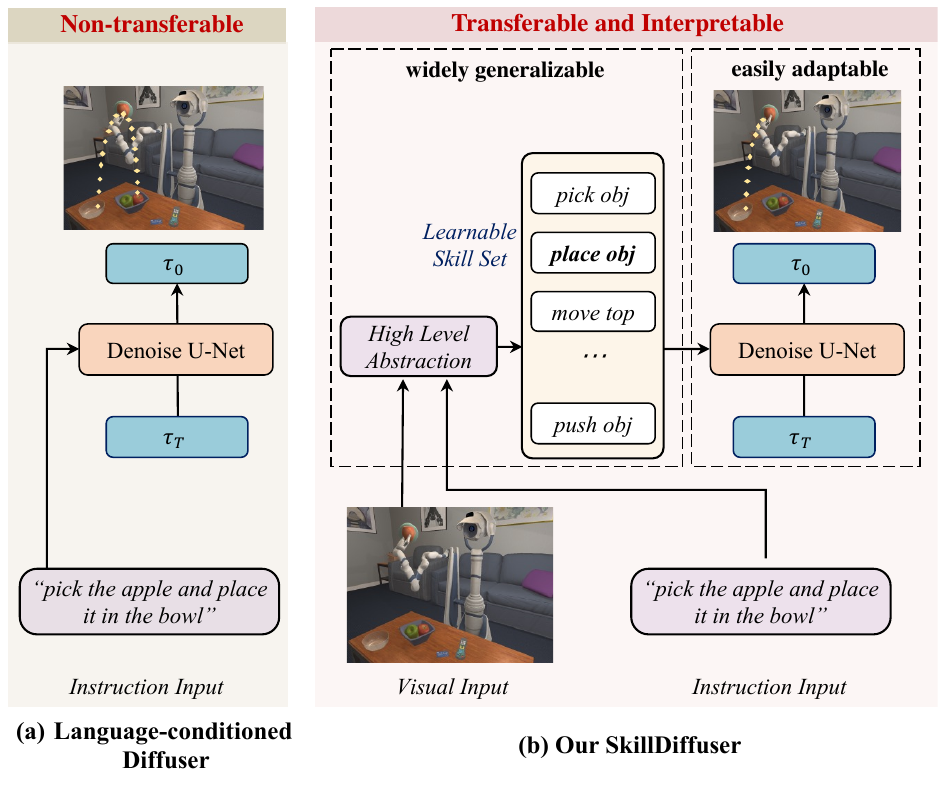}
\vspace{-8pt}
\caption{\textbf{Comparison of \alias and previous language conditioned diffusers.} \alias utilizes high-level abstraction to translate visual observations and language instructions into human understandable skills with language grounding.
It then enables the low-level diffusion model condition on these skills, not only improving the execution performance of multi-step composition tasks but greatly enhancing the generalization and adaptability of the framework.}
% effectively improving the execution performance of complex, abstract, and multi-step composite instructions through long-horizon task decomposing.}
\vspace{-15pt}
\label{fig:teaser} 
\end{figure}

% 打一个语言到motion的生成

% 用diffusion model 替换decision transformer旨在通用，只有inverse dynamics model不同，别的相同

% 用classifier-free diffusion model, 用skill做condition

% Recent research has shown that diffusion models can generate high-quality samples and train more stably compared to previous generative models. Diffusion models have been used to improve reinforcement learning in several ways, such as generating trajectories, representing policies, and synthesizing data. However, diffusion models struggle to generate long, coherent trajectories for complex tasks that require many steps to complete, especially when the task description is abstract. 

Recent research~\cite{ho2020denoising,janner2022planning,du2023learning,dhariwal2021diffusion} has demonstrated diffusion models'
superior generative capabilities compared to previous models
that 
% diffusion models outperform previous generative models in generating high-quality samples and exhibit stronger training stability. It has been proven that these diffusion models are instrumental in enhancing 
help enhance reinforcement learning across various dimensions, including the generation of action trajectories~\cite{janner2022planning,decisiondiffuser}, policy representation~\cite{yang2023policy,chi2023diffusionpolicy}, and data synthesis~\cite{he2023diffusion,liang2023adaptdiffuser}. 
However, their ability to generate coherent trajectories for intricate tasks still poses challenges in terms of performance and generalizability, as these tasks often require the fulfillment of abstract instructions that consist of numerous coordination-intensive sequential steps.

% Currently, existing methods struggle to infer high-level objectives and key steps purely from language descriptions when no explicit sub-goals are defined.
% when there are no explicit sub-goals, existing methods find it difficult to fully infer high-level goals and key steps from language descriptions.

Previous approaches~\cite{he2023diffusion, decisiondiffuser}, such as Decision Diffuser, aim to tackle this challenge by decomposing complex tasks into simpler sub-skills, organized within a predefined skill library. These methods rely on conditioning the diffusion model with one-hot skill representations to generate trajectories for each of these sub-skills.
Nevertheless, these techniques encounter difficulties in attempting to autonomously learn end-to-end from a wide range of datasets, which hinders their scalability and ability to achieve end-to-end learning~\cite{saycan}.
In addition, without explicitly learning reusable skills, models cannot capture intricate inter-step dependencies and constraints, yielding fragmented and illogical trajectories. Decomposing ambiguous instructions into learnable sub-goals and skills can better enable models to follow logical step sequences, respect task structures, and transfer common procedures (\eg reusable skill abstraction and adaptable skill-based diffusion) between different tasks. This skill-centric paradigm paves the way for diffusion models that can interpret and execute elaborate, abstract instructions necessitating numerous sequential steps.

Inspired by the above observations, we propose \alias, a hierarchical planning framework unifying high-level skill learning with low-level conditional diffusion-based execution. As shown in Fig.~\ref{fig:teaser}, compared with previous language-conditioned diffusion policies, \alias is able to interpret and execute complex instructions end-to-end with higher transferability.
%
% To enable diffusion models to interpret and execute complex instructions end-to-end, we propose \alias, a hierarchical planning framework unifying high-level skill learning with low-level conditional diffusion model-based planning. A comparison between our method and the previous language-conditioned diffuser is shown in Fig.~\ref{fig:teaser}. 
\alias induces interpretable sub-latent goals by learning reusable skills tailored to the task instructions. The framework conditions a diffusion model on these learned skills to generate customized, coherent trajectories aligned with the overall objectives. By integrating hierarchical skill decomposition with conditional trajectory generation, \alias achieves consistent, skill-oriented behavior without relying on a predefined skill library. Moreover, \alias is designed to operate solely on visual observations, eliminating the need for robot proprioception (\ie fully observed states). This end-to-end methodology, featuring learnable skills, enables \alias to execute abstract instructions across a variety of tasks efficiently.

% By abstracting complex tasks into skills (sub-goals), \alias enables the diffusion model to work in concert with these sub-goals, ensuring adaptability and flexibility.

% a novel hierarchical planning framework integrating the sequence generation strengths of diffusion models with interpretable skill representations.
% SkillDiffuser decomposes ambitions instructions into sub-goal abstractions and learnable skills to...

\alias works as follows. 
\textbf{1)} A skill predictor with vector quantization~\cite{vq-vae} is used for high-level skill learning to distill tasks into discrete and interpretable skills. Rather than forecasting skill durations, we adopt a fixed prediction horizon -- predicting skills at regular intervals. This horizon-based discretization process seamlessly integrates visual and linguistic inputs into a cohesive skill set guiding the low-level diffusion model.
\textbf{2)} For skill-based trajectory generation, we utilize a classifier-free diffusion model as policy, with skills directly embedded as guidance. This setting allows for generating multi-modal state trajectories aligned with skill specifications while avoiding overfitting to a closed dataset. 
\textbf{3)} To enable action inference from predicted states, we train an inverse dynamics network to decode motions between two consecutive frames generated. By separating state prediction from motion decoding, \alias yields a fully adaptive framework for directing diverse embodiments via transferable state-space plans.
% can generalize state forecasts to new environments by simply adapting the inverse dynamics, as the only environment-specific component, to new embodiments. 
% The model's state predictions are refined by an inverse dynamics model, which deciphers actions to transition between generated states, reinforcing the generalization capability of the SkillDiffuser across multiple tasks. This generalization is made possible by maintaining a modular approach, where the inverse dynamics model is the only component that requires adaptation when introduced to new task environments.

We evaluate the model's performance in both skill learning and multi-task planning on the \lorl~\cite{lorl} Sawyer and \metaworld~\cite{yu2020meta} datasets, with crucial experimental settings by considering robotic agents in real-world scenarios must operate with limited state information, chiefly visual observations.
% Our experiments are conducted on 
% multiple datasets, including 
% the \lorl~\cite{lorl} Sawyer and \metaworld~\cite{yu2020meta} datasets, to assess the model's performance both in skill learning and multi-task planning.
% These tests are important and impactful, especially considering real-world robotic scenarios where agents must operate with limited state information, chiefly visual observations. 
% \alias's state-of-the-art performance on these benchmarks demonstrates its ability to learn and execute multifaceted skills.
Furthermore, we present success rates on unseen compositional tasks, the reusability and visualizations of learned skills to illustrate the model does have the ability to abstract high-level skills that are not only effective but also understandable to humans, bringing us closer to intelligent agents that acquire skills in a direct manner.

% In conclusion, \alias's introduction enables the hierarchical decomposition and execution of complex tasks by diffusion models through interpretable skill learning end-to-end.
Our contributions are three-fold: 
\textbf{1)} We propose an end-to-end hierarchical planning framework via skill learning for sub-goal abstraction;
\textbf{2)} We adopt a classifier-free diffusion model conditioned on learned skills to generate skill-oriented transferable state trajectories; 
\textbf{3)} We demonstrate state-of-the-art performance on complex benchmarks and provide interpretable visualizations of human-understandable skill representations.

\section{Related Works}
\label{sec:relate_work}
\subsection{Imitation Learning and Multi-task Learning}
Imitation learning (IL) has evolved from foundational behavioral cloning to sophisticated multi-task learning frameworks. With traditional approaches relying on supervised learning from expert demonstrations~\cite{pomerleau1991bc,ross2010efficient,ross2011reduction}, recent advancements have shifted towards learning the reward~\cite{ho2016generative} or Q-function~\cite{garg2021iq} from expert data, enhancing the ability to mimic complex behaviors. A new challenge lies in multi-task IL~\cite{singh2020scalable}, where imitators are trained across varied tasks, aiming for generalization to new scenarios with task specifications ranging from vector states~\cite{nair2018visual} to vision and language descriptions~\cite{lisa,duan2017one,xu2018neural,finn2017one}.

Multi-task learning approaches often leverage shared representations to learn a spectrum of tasks simultaneously, enhancing the flexibility and efficiency of the learning process. The Meta-World benchmark~\cite{yu2020meta} assesses multi-task and meta reinforcement learning, highlighting the need for algorithms capable of rapid adaptation. Building on this, the Prompting Decision Transformer~\cite{xu2022prompting} showcases few-shot policy generalization using task-specific prompts. And diffusion policy has also been explored in multi-task settings~\cite{he2023diffusion}, which shows proficiency in generating diverse behaviors across tasks.
However, unlike methods that leverage state inputs~\cite{he2023diffusion,xu2022prompting} or access robot proprioception~\cite{r3m}, \alias uses raw visual inputs only.

% While these works represent significant strides in multi-task learning, the reliance on state-based inputs rather than pure visual inputs is a notable limitation, pointing to an important direction for future research.

% Some of current researches~\cite{mandlekar2021matters,rashidinejad2021bridging,kumar2022should} try to bridge the gap between offline reinforcement learning (RL) and IL. For example, Xu \etal~\cite{xu2022discriminator} explored discriminator-weighted techniques for reinforcement learning from sub-optimal demonstrations. Liu \etal~\cite{liu2021curriculum} introduced a structured approach to offline imitation learning, where a curriculum is designed to facilitate learning in a more organized and effective manner. Moreover, in the most recent work, MAHALO~\cite{mahalo} was proposed, a unified framework that encompassed offline imitation learning, imitation learning from observations, and offline reinforcement learning, indicating a trend towards a more holistic view of learning from demonstrations.

% In conclusion, these integrative approaches are setting the stage for robust, versatile imitation learning applications in offline reinforcement learning, with a growing focus on language as a natural conduit for task specifications.

\subsection{Skill Discovery and Hierarchical Learning}
Skill learning, the process by which robots acquire new abilities or refine existing ones, is gaining increasing attention due to its pivotal role in enabling autonomous systems to adapt to new tasks, improve the performance over time, and interact naturally with humans and complex environments.
Traditionally, this domain was influenced by hand-crafted features and expert demonstrations~\cite{niekum2015learning}. 

With the development of deep learning,
% Deep learning has sparked the development of innovative methodologies. For example, 
Eysenbach~\etal~\cite{eysenbach2018diversity} and A. Sharma~\etal~\cite{sharma2019dynamics} investigated skill discovery in learning methods, achieving policies conditioned on learned latent variables and maintaining consistent skill codes throughout trajectories.
In the domain of skill learning through language instructions, LISA~\cite{lisa} stands out by sampling multiple skills per trajectory, integrating language conditioning in a unique manner. 

Our \alias follows this way to extract sub-skills of each instruction at the higher-level. But differently, \alias employs an adaptable diffusion policy at the lower-level to condition on different sub-skills to generate different actions, which formulates a creative hierarchical planning framework also advancing research on hierarchical techniques of reinforcement learning~\cite{zhang2021hierarchical,nachum2018data,li2019sub}.

% These contributions enrich the field of skill discovery and hierarchical learning, offering diverse perspectives and methodologies.

% Zhang \etal harnessed virtual reality headsets for robot teleoperation and subsequently leveraged deep imitation learning to capture demonstrated skills~\cite{zhang2018deep}.

\subsection{Planning with Diffusion Model}
% In the past few year, the realms of robotic control and decision making have seen significant advancements with the integration of machine learning~\cite{chen2021decision,janner2021offline}. One of the recent breakthroughs in this domain is the use of diffusion model~\cite{ho2020denoising} that is originally designed for image synthesis~\cite{ho2020denoising} and has shown promising results in various generative applications.

Diffusion models~\cite{ho2020denoising} make great breakthroughs in image synthesis~\cite{ho2020denoising} recent years and has shown promising results in various generative applications. A seminal work that performs planning with diffusion directly is Diffuser~\cite{janner2022planning}, which laid the groundwork for using diffusion models in behavior synthesis. Then, a branch of this kind planning methods achieved state-of-the-art performance in a variety of decision tasks~\cite{ni2023metadiffuser,carvalho2023motion,du2023learning,liang2023adaptdiffuser,chi2023diffusionpolicy}. Among them, the work done by Chi~\etal in~\cite{chi2023diffusionpolicy} introduces the concept of learning the gradient of the action-distribution score function and iteratively optimizing it, demonstrating its significant potential in visuomotor policy learning. These works have further extended this direction and strengthened the versatility and generalization of diffusion-based planners.
% The representative work~\cite{dhariwal2021diffusion} showcased diffusion models have the ability to outperform Generative Adversarial Networks in generating high-fidelity images, providing a strong foundation for their potential in other domains, including robotic control and planning.

Our approach is inspired by classifier-free diffusion guidance~\cite{ho2021classifier}, which offers a significant advantage over classifier-guided models~\cite{dhariwal2021diffusion}. By adopting the classifier-free approach, we can circumvent the challenges associated with training a reward model and $Q$-function, which are particularly cumbersome in many real-world planning scenarios of which the complexity is very high. 
Recent studies have also extended this direction, which use conditional diffusion models to generate customized trajectories. Decision Diffuser~\cite{decisiondiffuser} is an example which is designed to generate trajectories for a predefined skill library. 
However, unlike our method, it can't autonomously learn skill abstractions in an end-to-end fashion, which makes it difficult to scale to more tasks. This highlights the necessity for diffusion models with dynamic, learnable skill abstractions, facilitating complex instruction execution.

\section{Preliminary}
\subsection{Planning with Diffusion over States}
% \vspace{-2pt}
% \subsection{Problem Formulation}
As introduced in previous works~\cite{janner2022planning,decisiondiffuser}, diffusion model is a promising tool to address the problem of planning in reinforcement learning which is cast as a Markov Decision Process (MDP)~\cite{puterman1994markov}. Within the MDP framework $\mathcal{M}=(\mathcal{S}, \mathcal{A}, \mathcal{T}, \mathcal{R}, \gamma)$, the planning policy aims to identify an optimal action sequence $\ba_{0:T}^*$ that maximizes the expected cumulative rewards over a finite time horizon $T$, governed by the state transition dynamics $\mathcal{T}$ and reward function $\mathcal{R}$. $\mathcal{S}$ is the state space and $\mathcal{A}$ is the action space.

By treating the state trajectory as sequence data $\btau$, with sequence modeling, diffusion probabilistic models conceptualize planning as an iterative denoising process. The model progressively refines trajectories by reversing a forward diffusion process that is modeled as a Gaussian process, whereby noise is incrementally added to the data, denoted as $p_\theta(\btau^{i-1} \mid \btau^i)$. Training involves minimizing the ELBO of the data's negative log-likelihood, similar to variational Bayesian inference, with the optimization objective:
% \vspace{-2pt}
% \begin{equation}
% % \small
% p_\theta\left(\btau^0\right)=\int p\left(\btau^N\right) \prod_{i=1}^N p_\theta\left(\btau^{i-1} \mid \btau^i\right) \mathrm{d} \btau^{1: N}
% % \vspace{-2pt}
% \end{equation}
\vspace{-8pt}
\begin{equation}
% \small
\theta^*=\arg \min _\theta-\mathbb{E}_{\btau^0}\left[\log p_\theta\left(\btau^0\right)\right],
\label{eq:theta_optim}
\vspace{-5pt}
\end{equation}
where $p\left(\btau^N\right)$ is a standard normal distribution and $\btau^{0}$ denotes noiseless sequence data.

For practical implementation, a simplified surrogate loss function is proposed in~\cite{ho2020denoising}, focusing on predicting the Gaussian mean of the reverse diffusion step:
\vspace{-3pt}
\begin{equation}
% \small
\label{eq:diff_loss}
    \mathcal{L}_{\text{denoise}}(\theta) = \mathbb{E}_{i, \btau^{0} \sim q, \epsilon \sim \mathcal{N}}[||\epsilon - \epsilon_{\theta}(\btau^i, i)||^{2}].
\end{equation}

\subsection{Classifier-free Diffusion Guidance}
On the basis of unconditioned diffusion-based method, in the realm of offline reinforcement learning, a critical endeavor is to generate trajectories with the highest reward-to-go. With the flourishing development of conditional diffusion models~\cite{dhariwal2021diffusion}, classifier-guided approaches embark on this by infusing specific trajectory information (encoded in $\boldsymbol{y}(\btau)$), such as the return $\mathcal{J}(\btau^0)$ or designated constraints, into the diffusion process:
\vspace{-5pt}
\begin{equation}
\label{eq:diff_plan}
    q(\btau^{i} | \btau^{i-1}), \;\;\;\; p_{\theta}(\btau^{i-1}|\btau^{i}, \boldsymbol{y}(\btau)).
    \vspace{-3pt}
\end{equation}
With assumptions specified in~\cite{feller2015theory}, we have
\vspace{-3pt}
\begin{equation}
      \btau^{i-1} \sim \mathcal{N}(\mu_{\theta} + \alpha\Sigma \nabla_{\btau} \log p\left(\boldsymbol{y}(\btau^{i})\right), \Sigma),
\vspace{-5pt}
\end{equation}
where $\alpha$ is a hyperparameter that adjusts the guidance strength, $\Sigma$ is the specified covariance of the noise and $\mu_{\theta}$ is the learned mean value of noise in unconditional diffusion.

However, the classifier-guided diffusion model requires an accurate estimation of guidance gradient based on the trajectory classifier $\boldsymbol{y}(\btau)$, which may not be feasible and need to introduce a separate dynamic programming procedure to estimate a $Q$-function in the training process. 

Thus, classifier-free guidance offer an alternative, which augments the trajectory generation process with a guidance signal that amplifies the features of high-reward or optimal characteristics, \ie $\boldsymbol{y}(\btau)$, that are implicitly present in the data. Mathematically, the noise to add during the reverse denoising process is:
\vspace{-5pt}
\begin{equation}
    \hat{\epsilon} = \epsilon_\theta(\btau^i, \varnothing, i) + \omega(\epsilon_\theta(\btau^i, \boldsymbol{y}, i) - \epsilon_\theta(\btau^i, \varnothing, i)),
    \label{eq:hat_epsilon}
    \vspace{-5pt}
\end{equation}
where $\omega$ is the guidance scale, and $\varnothing$ represents the absence of guidance. Setting $\omega = 0$ removes the classifier-free guidance towards an unconditional diffusion model, while a large value of $\omega$ strengthens the influence of the conditional information during trajectory generation.

Also, the loss function to minimize can be rewritten as,
\vspace{-5pt}
\begin{equation}
    \mathcal{L}_{\mathit{diff}}(\theta) = \mathbb{E}_{i, \btau, \epsilon}\left[\left\|\epsilon - \epsilon_\theta\left(\btau^i, (1 - \beta) \boldsymbol{y}(\btau^i) + \beta \varnothing, i\right)\right\|^2\right],
    \label{eq:loss_diff}
\end{equation}
where $\beta$ is a hyperparameter that controls the probability of dropping specific condition $\boldsymbol{y}(\btau)$, enhancing sample diversity while maintaining context relevance.

After training the noise prediction model $\epsilon_\theta$ with the above $\mathcal{L}(\theta)$, the trajectory is sampled starting from Gaussian noise and progressively denoised with modified $\hat{\epsilon}$ through Eq.~\ref{eq:hat_epsilon}, employing re-parameterization technique.

In summary, through classifier-free guidance, we can modulate the trajectory sampling process to require the generated trajectories more aligned with the desired features represented by $\boldsymbol{y}(\btau)$. This process iteratively applies the conditioned noise model to refine the target trajectories that contain future states satisfying the constraints.

\section{Methodology}

\subsection{Overview}
\begin{figure}[tb]
\centering 
\includegraphics[width=0.85\linewidth]{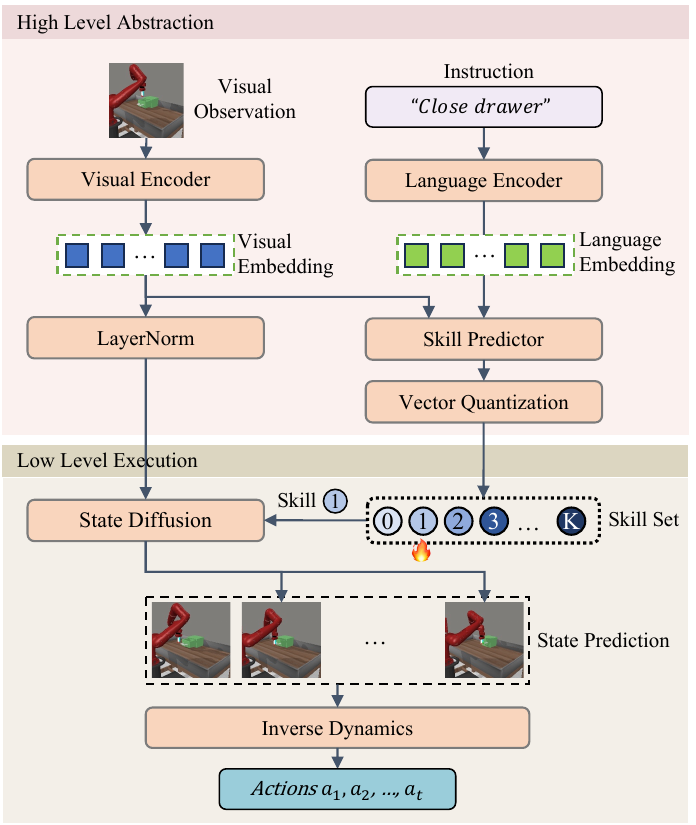}
\vspace{-5pt}
\caption{\textbf{Overall framework of \alias.} It's a hierarchical planning model that leverages the cooperation of interpretable skill abstractions at the higher level and a skill conditioned diffusion model at the lower level for task execution in a multi-task learning environment. The high-level skill abstraction is achieved through a skill predictor and a vector quantization operation, generating sub-goals (skill set) that the diffusion model employs to determine the appropriate future states. Future states are converted to actions using an inverse dynamics model. This unique fusion enables a consistent underlying planner across different tasks, with the variation only in the inverse dynamics model.}
\vspace{-18pt}
\label{fig:main_frame} 
\end{figure}

Building upon the motivations discussed above, we present \alias, an advanced methodological framework for robust multi-task learning across various robots. This dynamic approach leverages the cooperation of skill learning at the higher level and a conditional diffusion model at the lower level. An overall framework is illustrated in Fig.~\ref{fig:main_frame}. Notably, we leverage language-grounded representations to abstract skills, thereby rendering the execution of tasks through our diffusion policy both interpretable and comprehensible to humans. 
% This approach significantly enhances the human understandability of the model's decision-making process.

\begin{figure*}[tb]
\vspace{-5pt}
\centering 
\includegraphics[width=0.85\linewidth]{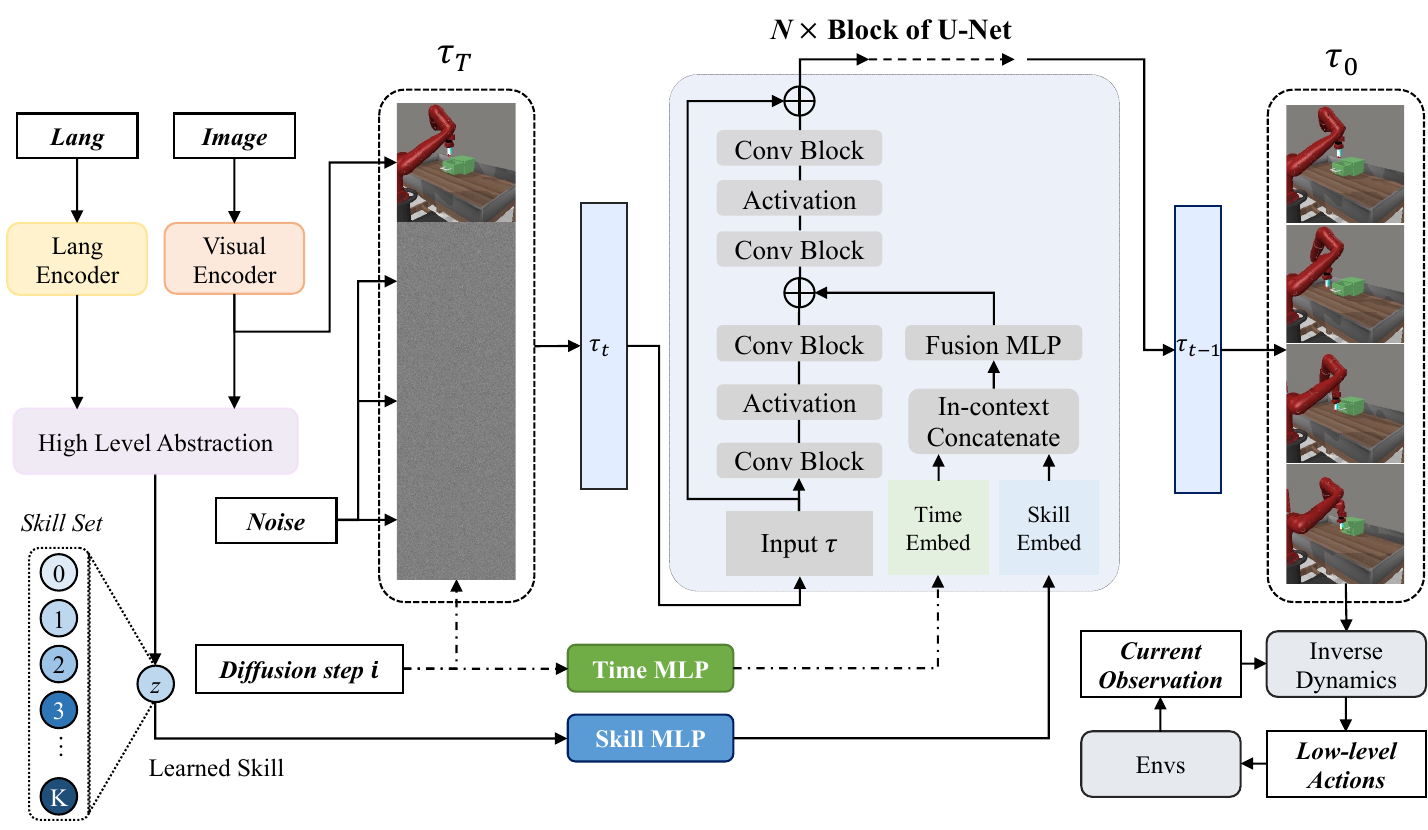}
\vspace{-7pt}
\caption{\textbf{\alias's low level skill-conditioned diffusion planning model.} Notably, while the schematic here employs images to represent visual features for illustrative purposes, in actual implementation, both the input to and the sampling output of the diffusion model are state embeddings. The current observation is also the feature embedding of current visual observation.}
\vspace{-18pt}
\label{fig:detail} 
\end{figure*}

\subsection{High Level Interpretable Skill Abstractions}
In our \alias framework, the high level interpretable skill abstraction module plays a crucial role in understanding and executing complex tasks. However, given the multi-task environments we consider, each task with only a single language instruction may be broken down into a sequence of sub-tasks or skills, which are not explicitly delineated within the instruction itself.
Furthermore, suppose the offline training dataset is denoted as \(\mathcal{D}\), it consists of trajectories derived from a sub-optimal policy for various tasks. A trajectory $\btau = (l, \{\vct{i}_t, \vct{a}_t\}_{t=1}^T)$ includes language description $l$ and a sequence of image observations and actions $(\vct{i}_t, \ba_t)$ over time steps $T$, with no reward labels attached. But the trajectories do not indicate the boundaries between sub-tasks as well, which thus requires the proposed methods capable of segmenting and interpreting the tasks into sub-goals in an unsupervised manner.

To address this challenge, we build a skill abstraction component upon a horizon-based skill predictor adapted from GPT-2~\cite{gpt2}. It is designed to parse and decompose tasks by fusing visual input and natural language instructions, alongside a Vector Quantization (VQ) sub-module that discretizes the learned skills into a skill set. The specifics of this component are illustrated below.

Firstly, as we utilize only images as robot state information, we transform the images into latent space features with a fixed image encoder (\eg R3M~\cite{r3m}).
% before images are processed by the skill predictor. 
For convenience, we denote the image encoder as \( \Phi_{im}: \mathcal{I} \rightarrow \mathcal{R}^I \), where \( \mathcal{I} \) represents the space of input images and \( \mathcal{R}^I \), the resultant feature space, serves to condense the visual information into a form that is conducive for high-level semantics. Simultaneously, we use a language encoder to pre-process the natural language instructions, which is formalized as \( \Phi_{lang}: L \rightarrow \mathcal{R}^L \), with \( L \) being the space of language instructions and \( \mathcal{R}^L \) the language feature space. The outputs of both encoders are then fed into the skill predictor, which operates to integrate these two modalities.

The skill predicting process is as follows: An image $\vct{i}_t \in \mathcal{I}$ at time step \( t \) is encoded into a visual embedding $\vct{s}_t = \Phi_{im}(\vct{i}_t)$. This embedding  $\vct{s_t}$ is then input to the skill predictor, along with the language instruction $l \in L$, through the language encoder's output \( \vct{l}_t = \Phi_{lang}(l) \). The skill predictor, represented as $f: \mathcal{R}^I \times \mathcal{R}^L \rightarrow \mathcal{C}$, generates a skill code $\vct{\tilde{z}}$ by $\vct{\tilde{z}} = f(\vct{s}_t, \vct{l}_t)$ that encapsulates task's requirements interpreted from the visual and language inputs.

After that, Vector Quantization~\cite{vq-vae} operation \(\mathbf{q(\cdot)}\) is taken to transform these predicted skills into a low-dimensional discrete set $\mathcal{C}$. The discrete skill set contains $K$ skill embeddings $\left\{\vct{z}^1, \vct{z}^2, \ldots, \vct{z}^{K}\right\}$ which represent different potential skills. VQ operation is achieved by mapping the latent $\tilde{\vct{z}}$ to its closest entry of the skill set with skill vectors updated to be the moving average of the embeddings $\vct{z}$ closest to them, same as~\cite{vq-vae}. This process is as follows,
\vspace{-2pt}
\begin{equation}
    \tilde{\vct{z}} = f\left(\Phi_{im}(\vct{i}_t), \Phi_{lang}(l)\right),
    \label{eq:cont_z}
    \vspace{-12pt}
\end{equation}
\begin{equation}
% \vspace{-5pt}
    \vct{z} = \mathbf{q}(\tilde{\vct{z}}) = \argmin _{\vct{z}^k \in \mathcal{C}}\|\tilde{\vct{z}}- \vct{z}^k\|_{2}.
    \label{eq:vq_z}
    \vspace{-2pt}
\end{equation}

VQ enforces each learnt skill $\vct{z}$ to lie in $\mathcal{C}$,  which is equal to learning $K$ prototypes for the language embeddings utilizing $k$-means~\cite{kmeans} algorithm. This acts as a bottleneck, restricting the flow of language information and aiding in the learning of discrete skill codes.
The back-propagation through the non-differentiable quantization is achieved by a straight-through gradient estimator, which simply copies the gradients and enables model to be trained end-to-end.

In our approach, we apply a consistent skill code $\vct{z}$ across a defined number of time steps, termed the horizon. This consistent application across multiple horizons adeptly addresses the challenge of varying sub-task durations without altering the horizon itself. Consequently, this strategy not only preserves the flexibility required for diverse task execution but also maintains the model's architectural stability by avoiding horizon-induced structural changes.

% We make each selected skill code $\vct{z}$ maintain across a designated number of time steps, called as horizon, subtly navigates the challenge of various sub-task durations by simply retaining the same skill code across multiple horizons. This strategy not only preserves the flexibility required for diverse task execution but also maintains the consistency of the model's architecture.
% denoted as $H$, the horizon. By sustaining the skill code throughout this horizon, it functions as a temporal abstraction, equal to the ``option'' in decision-making processes. This design, anchored in the concept of horizons, 

Importantly, the discrete nature of the learned skill codes enhances the interpretability and controllability of the system's behavior, as each skill code is associated with some human-understandable language phrases. An example is depicted in Fig~\ref{fig:heatmap_MT10}. Through this method, \alias can not only learn to perform tasks based on language instructions but also achieve them in a human-interpretable way, allowing for a deeper understanding and control of the decision-making process of specific embodied agents.

\subsection{Low Level Skill-conditioned Diffusion Planning}

As highlighted in Section~\ref{sec:intro}, while existing approaches like Decision Diffuser~\cite{decisiondiffuser} have introduced conditional diffusion models constrained by skills, their capability is limited to generating trajectories that meet only predefined skills. Consequently, these models fall short of realizing a diffusion model capable of conditioning on an arbitrary learned skill.
To overcome these constraints and enable diffusion models to plan over a learned continuous spectrum of skills, we propose an approach that leverages the classifier-free guided diffusion model with the skills embedded.

\alias begins by employing a diffusion model to operate over the continuous skill space $\mathcal{Z}$ learned during high-level skill abstraction. We employ a U-Net to serve as the noise prediction model $\boldsymbol{\epsilon}_{\theta}(\cdot)$ and guide it by in-context conditioning. More specifically, we firstly utilize a skill MLP (similar to point-wise feed-forward network~\cite{transformer}), denoted as $\Lambda$, to align skill features with denoising model. After that, we fuse the skill embeddings $\Lambda(\vct{z})$ into the state features throughout the residual blocks of U-Net. Details are depicted in Fig.~\ref{fig:detail}. In this way, we make the diffusion model contextually modulate the influence of skill embeddings at each step. And following Eq.~\ref{eq:hat_epsilon}, we can synthesize future states attending to these given skills. This is a significant shift from previous static conditioning framework to a more dynamic and adaptive trajectory generation process.

Moreover, following previous works~\cite{decisiondiffuser,liang2023adaptdiffuser,du2023learning}, we adopt a state-only diffusion model, which eschews the direct generation of actions in diffusion model.
% to facilitate more generalizable state predictions. 
And instead, we utilize another MLP, denoted as $\Psi(\cdot)$, to perform inverse dynamics after state generation to infer feasible actions that can achieve transitions between two continuous states. Additionally, we integrate observations from the current frame to provide more detailed information for motion prediction and achieve closed-loop control. Mathematically, it is:
\vspace{-5pt}
\begin{equation}
    \ba_t = \Psi([\tilde{\bs}_t, \tilde{\bs}_{t+1}], \vct{i}_t) \quad \text{for} \quad t=0, \ldots, T-1,
    \label{eq:inv}
    \vspace{-5pt}
\end{equation}
where $\tilde{\bs}_t$ and $\tilde{\bs}_{t+1}$ are consecutive observation embeddings within $\tilde{\btau}$ generated by diffusion model, $\vct{i}_t$ is the current observation and $\ba_t$ is the inferred action.

As the resulting action sequence $\{\ba_0, \ba_1, \ldots, \ba_{T-1}\}$ derived from the generated states, it encapsulates the skills to execute the tasks, which allows for remarkable adaptability across multiple tasks. And when faced with a new task, we are only required to change the inverse dynamics model $\Psi(\cdot)$ specific to the new task's kinematics, with the architecture and parameters of diffusion model unchanged. Such modularity ensures the generative capabilities of \alias are not task-specific but can be leveraged across a diverse range of tasks with varying dynamics.
% which mirrors the complex skill interplay and adaptability seen in human learning. 
A schematic diagram of the low level module is shown in Fig.~\ref{fig:detail}.

\subsection{Training the \alias}
\label{sec:train_process}
We engineered a threefold loss function for \alias.
% with each meticulously designed and straightforward to meet the distinct needs of respective modules within \alias.
Firstly, for the inverse dynamics model which is task-specific, we employ a behavior cloning loss~\cite{torabi2018behavioral} to train our inverse MLP emulating expert actions from observed state transitions. This loss, denoted as \( \mathcal{L}_{inv} \), is formulated as:
\vspace{-5pt}
\begin{equation}
% \small
    \mathcal{L}_{inv} = \mathbb{E}_{\btau \sim \mathcal{D}}\left[ \left\| \ba - \Psi(\tilde{\bs}, \tilde{\bs}', \vct{i}) \right\|_{2}^{2} \right],
    \label{eq:loss_inv}
    \vspace{-5pt}
\end{equation}
where the notations are similar to Eq.~\ref{eq:inv}.

Correspondingly, the other parts including both skill abstraction and low level execution are task-agnostic. For the high-level skill abstraction module, we apply a vector quantization (VQ) loss to refine the skill predictor. This VQ loss, \( \mathcal{L}_{\mathrm{VQ}} \), ensures the embeddings produced by the skill predictor closely match the skill set vectors, thereby improving the interpretability and consistency of the skill representations. Inspired by VQ-VAE~\cite{vq-vae}, we formulate it as:
\vspace{-5pt}
\begin{equation}
% \small
\mathcal{L}_{\mathrm{VQ}}=\mathbb{E}_{\btau}\left[\|\mathbf{q}(\tilde{\vct{z}})-\tilde{\vct{z}}\|_{2}^{2}\right],
    \label{eq:loss_vq}
    \vspace{-5pt}
\end{equation}
where $\tilde{\vct{z}}$ follows Eq.~\ref{eq:cont_z}.
% and $\operatorname{sg}[\cdot]$ is stop-gradient operation.

Lastly, for the low-level state-only skill-conditioned diffusion execution, we incorporate a diffusion loss $\mathcal{L}_{\mathit{diff}}$ as per Eq.~\ref{eq:loss_diff}, ensuring our model's state predictions are in line with both the skill guidance and temporal dynamics observed in expert demonstrations. Here, we take $\boldsymbol{y}(\btau) = \Lambda(\vct{z})$ with $\vct z$ derived from Eq.~\ref{eq:cont_z} and Eq.~\ref{eq:vq_z}.

To be noted, we train our \alias with two optimizers, one for inverse dynamics model with $\mathcal{L}_{inv}$ and the other for overall parameters of high-level skill abstraction and low-level planning with $\mathcal{L}_{\mathrm{VQ}} + \lambda \mathcal{L}_{\mathit{diff}}$, where $\lambda$ is a loss weight. This carefully constructed loss architecture enables \alias to excel in a multi-task environment, generalizing across tasks by simply substituting the inverse dynamics model $\Psi(\cdot)$ specific to each new task's requirements.

Additionally, we utilize a pre-trained distilBERT~\cite{distilbert} as our language encoder, adopting the configuration consistent with \lorl~\cite{lorl} and LISA~\cite{lisa}, while freezing its parameters to guarantee stability in language understanding.
% Additionally, the language encoder is achieved by leveraging a pre-trained distilBERT~\cite{distilbert} with parameters frozen to ensure stability in the language understanding component. 
And we employ diverse settings to serve as the visual encoder to ensure fair comparison in different datasets. We elaborate the details in corresponding parts of Sec.~\ref{sec:experiment}. More training details are shown in Appendix~\ref{append:train_detail} and we also provide some pseudo-code of our algorithm in Appendix~\ref{append:pseudo}.
\begin{table*}[tb]
    \centering
    \resizebox{0.86\linewidth}{!}{
	\begin{tabular}{l c c c c c c}
        \toprule
		\textbf{Task Instruction} & \textbf{Random} & \textbf{LCBC~\cite{LCBC}} & \textbf{LCRL~\cite{LCRL}} & \textbf{Lang DT~\cite{lisa}} & \textbf{LISA~\cite{lisa}} & \textbf{\alias} \\
        \midrule
		close drawer & 52\% & 50\% & 58\% & 10\% & \textbf{100\%} & \textbf{95 \small{\raisebox{.5pt}{$\pm$ 3.2}}}\%\\
		open drawer & 14\% & 0\% & 8\% &\textbf{60\%} & 20\% & \textbf{55 \small{\raisebox{.5pt}{$\pm$ 13.3}}\%}\\
		turn faucet left & 24\% & 12\% & 13\% & 0\% & 0\% & \textbf{55 \small{\raisebox{.5pt}{$\pm$ 9.3}}\%}\\
		turn faucet right & 15\% & \textbf{31\%} & 0\% & 0\% & \textbf{30\%} & \textbf{25 \small{\raisebox{.5pt}{$\pm$ 4.4}}\%}\\
		move black mug right & 12\% & \textbf{73\%} & 0\% & 20\% & 60\% & 18 \small{\raisebox{.5pt}{$\pm$ 3.9}}\%\\
		move white mug down & 5\% & 6\% & 0\% & 0\% & \textbf{30\%} & 10 \small{\raisebox{.5pt}{$\pm$ 1.7}}\%\\
          \midrule
        \textbf{Average over tasks} & 20\% & 29\% & 13\% & 15\% & 40\% & \textbf{43 \small{\raisebox{.5pt}{$\pm$ 1.1}}\%}\\
	\bottomrule
    \end{tabular}}
    \vspace{-5pt}
    \caption{\textbf{Task-wise success rates on \lorl Sawyer Dataset.} We show our success rates compared to random policy, language-conditioned imitation learning (LCBC), language-conditioned Q-learning (LCRL), a flat non-hierarchical Decision Transformer (Lang-DT), and LISA. The results on each dataset are calculated over 3 seeds. \alias outperforms all other methods in terms of average performance over all tasks. Best methods and those within 6\% of the best are highlighted in \textbf{bold}.}
    \label{tab:lorel_res_1}
    \vspace{-15pt}
\end{table*}

\begin{table}[tb]
    \centering
	\small
% 	\vskip-3pt
% 	\tabcolsep 3pt
	% \vskip-5pt
	\begin{tabular}{lccc}
 \toprule
		\textbf{Rephrasal Type} & \textbf{Lang DT} & \textbf{LISA}~\cite{lisa} & \textbf{\alias} \\ 
  \midrule
		seen & 15 & 40 & \textbf{43.65} \scriptsize{\raisebox{1pt}{$\pm 4.7$}}\\
		unseen noun &  13.33 & 33.33 & \textbf{36.01} \scriptsize{\raisebox{1pt}{$\pm 6.3$}}\\
		unseen verb & 28.33 & 30 & \textbf{36.70} \scriptsize{\raisebox{1pt}{$\pm 9.5$}}\\
		unseen noun+verb & 6.7 & 20 & \textbf{42.02} \scriptsize{\raisebox{1pt}{$\pm 3.8$}}\\
		human provided & 26.98 & 27.35 & \textbf{40.16} \scriptsize{\raisebox{1pt}{$\pm 2.1$}}\\
      \midrule
        \multicolumn{1}{l}{\textbf{Average}} & 18.07 & 30.14 & \textbf{39.71} \hspace{.58cm} \\
		\bottomrule
    \end{tabular}
    \vspace{-8pt}
	\caption{\textbf{Rephrasal-wise success rates (in \%) on \lorl Sawyer.} Results of Lang DT, LISA and our \alias are shown here. The standard error is calculated over 3 random seeds.}
 \vspace{-15pt}
 \label{tab:lorel_res_2}
\end{table}

\section{Experiments}
\label{sec:experiment}
We first present a comprehensive evaluation on the \lorl Sawyer dataset, and then perform the ablation study compared on \metaworld benchmark to illustrate the efficiency of our method. Finally, we visualize the learned skills of our method both on \lorl and \metaworld MT10.
% to see a human-understandable skill-conditioned diffusion-based planning.
\subsection{Datasets}
\textbf{\lorl Sawyer Dataset}~\cite{lorl}
\label{sec:lorl_dataset}
which is abbreviated from Language-conditioned Offline Reward learning, is composed of \emph{pseudo-expert} trajectories or \textit{play data} gathered from an arbitrary reinforcement learning policy, annotated with post-hoc crowd-sourced language directives.
% Although these trajectories fulfill the given linguistic commands, they are not guaranteed to be the optimal paths.
% Moreover, collecting \textit{play data} in real-world scenarios is cost-effective~\cite{lynch2020learning}, thus, it's equally important for algorithms to demonstrate robustness when working with these datasets.
% The inherent randomness within these trajectories presents a significant challenge for behavior cloning applications.
The \lorl Sawyer dataset encompasses 50k trajectories, each with 20 steps, in a simulated Sawyer robot environment. We assess our approach using the same six tasks as the original paper~\cite{lorl} with paraphrased instructions under five different situations (\ie seen, unseen verb, unseen none, unseen verb + noun and human provided). This comes to a total of 77 instructions for all 6 tasks combined. More details about the dataset can be found in Appendix~\ref{append:lorl}.

\vspace{0.1cm}
\noindent \textbf{\metaworld Dataset}~\cite{yu2020meta}.
It is a comprehensive benchmark designed for evaluating multi-task and meta reinforcement learning algorithms. It introduces a comprehensive suite of 50 distinct robotic manipulation tasks, all located within a unified tabletop environment controlled by a simulated Sawyer arm. 
The Multi-Task 10 (MT10) within Meta-World is a subset comprising ten carefully selected tasks, balanced in terms of diversity and complexity. Details of the ten tasks can be found in Appendix~\ref{append:metaworld}.

\subsection{Evaluation Results on \lorl Sawyer Dataset}
% \textbf{Settings.}
% We first perform experiments on \lorl Sawyer dataset with only partially observed image data.
% The analysis is anchored on six different tasks with paraphrased instructions under five different situations (\ie unseen verb, unseen none, human provided, \etc as introduced in Sec.~\ref{sec:lorl_dataset}). 
% This deliberate choice facilitates a direct comparison of our method against established benchmarks.
\textbf{Baselines.} 
We employ random policy, language conditioned imitation learning (LCBC)~\cite{LCBC}, language conditioned Q-learning (LCRL)~\cite{LCRL}, Lang-DT (also known as Flat Baseline in~\cite{lisa}) and state-of-the-art skill-learning method LISA~\cite{lisa} as our baselines. We follow the same settings as \lorl~\cite{lorl} for the first three algorithms, and follow LISA~\cite{lisa} for the last two. 
The random policy serves as a baseline.
% to provide a fundamental comparison point for the algorithm effectiveness.
And LCBC mimics offline dataset behaviors based on instructions, aligning with previous works focusing on imitation learning to achieve language conditioned behavior. In contrast, LCRL employs reinforcement learning, labeling each episode's final state with language-instructed rewards.
Lang-DT plays as a non-hierarchical benchmark with a Causal Transformer~\cite{melnychuk2022causal},
% embeddings from a pre-trained DistillBERT~\cite{sanh2019distilbert}, 
% diverging from traditional future sum of returns. 
while LISA incorporates a dual-transformer structure, with one for skill prediction and the other for action planning. We do not compare with \lorl planner~\cite{lorl} as it uses MPC on a learned reward function relying on human annotations, while LISA and ours learn with trajectory data only.
% about task completion to get the sub-optimal nature, 
% achieving state-of-the-art performance in language instructed multi-task learning. 
% To ensure parity in model complexity, Lang-DT is equipped with a transformer network featuring two self-attention layers, while LISA's skill predictor and policy each utilize a single self-attention layer. 

\vspace{2pt}
\noindent \textbf{Results.}
To ensure fair comparison, our method, \alias, is designed to maintain a parameter count similar to baseline models. It employs the same visual and language encoder architecture as used in LISA~\cite{lisa} with meticulously matching of embedding dimensions and the number of heads across the layers of \alias.

The results are present in Tables~\ref{tab:lorel_res_1} and \ref{tab:lorel_res_2}, showing task-wise and rephrasal-wise success rates for \lorl, averaged over 10 runs with a 20-step time horizon. \alias, our approach, achieves the highest average performance in six different tasks, indicating its superior cross-task adaptability particularly when compared to similar approaches which yet are based on Decision Transformer~\cite{DT}, such as LISA~\cite{lisa}. Moreover, \alias consistently excels in all rephrased types for \lorl test tasks, outperforming LISA by 9.6\% on average. This demonstrates the model's robustness in handling varied skill representations, marking a notable advancement in skill-conditioned diffusion model.

\begin{table}[tb]
    \centering
	\small
% 	\vskip-3pt
% 	\tabcolsep 3pt
	% \vskip-5pt
 \resizebox{\linewidth}{!}{
	\begin{tabular}{lcccc}
 \toprule
		\textbf{Method} & \textbf{Lang DT} & \textbf{\lorl}~\cite{lorl} & \textbf{LISA}~\cite{lisa} & \textbf{\alias} \\ 
  \midrule
		\textbf{Success Rate} & 13.33 \scriptsize{\raisebox{1pt}{$\pm 1.3$}} & 18.18 \scriptsize{\raisebox{1pt}{$\pm 1.8$}} & 20.89 \scriptsize{\raisebox{1pt}{$\pm 0.6$}} & \textbf{25.21 \scriptsize{\raisebox{1pt}{$\pm 2.7$}}}\\
		\bottomrule
    \end{tabular}}
    \vspace{-8pt}
	\caption{\textbf{Performance on \lorl multi-step composition tasks.}}
 \vspace{-15pt}
 \label{tab:composition}
\end{table}

\subsection{Performance on \lorl Compositional Tasks}
\textbf{Settings.} We conduct experiments following the same settings of unseen composition tasks of LISA~\cite{lisa} with 12 composition instructions in Tab.~\ref{tab:composition}. Detailed instructions are listed in Appendix~\ref{append:composition} with such an example that ``open drawer and move black mug right''. We extend the max number of episode steps from customary 20 to 40, as LISA.

\vspace{2pt}
\noindent \textbf{Results.}
We observe \alias achieves 2x the performance of non-hierarchical baseline (i.e. w/o skill abstraction) and also improves about 25\% over LISA, highlighting its effectiveness. MPC-based LOReL planners are unable to perform as well in open scenarios like composition tasks.

\subsection{Ablation Study on \metaworld Dataset}
\textbf{Settings.} We conduct experiments on \metaworld MT10 benchmark with finely annotated instructions.
% a standard multi-task dataset which contains ten carefully selected tasks with sub-optimal expert data. 
We also use visual observations only. Details are in Appendix~\ref{append:metaworld}.

\vspace{0.1cm}
\noindent \textbf{Baselines.}
We evaluate our method against three baselines, all modified from existing models. The first, Flat R3M, is adapted from R3M~\cite{r3m} paper's planner. As the original one utilizes the first four terms of robot proprioception,
% (\ie proprio = 4 in R3M implementation), 
we eliminate them and make the planner focus exclusively on visual observations. The second baseline is a variant of our \alias, lacking the high-level skill abstraction module but retaining the low-level conditional diffusion-based planner, to assess the impact of skill abstraction. This version integrates a two-layer MLP to predict options from visual and language inputs, functioning as a language-conditioned diffusion planner.
% thus preserving the original setting's core elements. 
The last baseline is our re-implemented LISA~\cite{lisa} for \metaworld Dataset to validate the efficiency of diffusion model.
To ensure fairness, we use R3M as the visual encoder for all of these methods on \metaworld.

\begin{table}[tb]
    \centering
	\small
% 	\vskip-3pt
% 	\tabcolsep 3pt
	% \vskip-5pt
 \resizebox{0.98\linewidth}{!}{
	\begin{tabular}{lcccc}
 \toprule
		\textbf{Method} & \textbf{Lang} & \textbf{Skill Set} & \textbf{Diffuser} & \textbf{Performance} \\ 
  \midrule
		Flat R3M~\cite{r3m} & \ding{55} & \ding{55} & \ding{55} & 13.3\% \\
		% Lang R3M &  \ding{51} & \ding{55} & \ding{55} &  \\
        LISA~\cite{lisa} & \ding{51} & \ding{51} & \ding{55} & 13.8\% \\
        Lang Diffuser & \ding{51} & \ding{55} & \ding{51} & 16.7\% \\
		\bestcell{\alias} & \bestcell{\ding{51}} & \bestcell{\ding{51}} & \bestcell{\ding{51}} & \bestcell{\textbf{23.3\%}} \\
      % \midrule
        % \multicolumn{1}{l}{\textbf{Average}} & 18.07 & 30.14 & \textbf{39.71} \hspace{.58cm} \\
		\bottomrule
    \end{tabular}}
    \vspace{-7pt}
	\caption{\textbf{Ablation study of language skill conditioning on \metaworld.} Results of Flat R3M, Language-condition diffuser, LISA and our \alias are shown here. All are averaged over 3 runs.}
 \vspace{-15pt}
 \label{tab:mateworld_res_1}
\end{table}

% \vspace{0.1cm}
\noindent \textbf{Ablation on Language Skill Conditioning.} Table \ref{tab:mateworld_res_1} indicates our \metaworld MT10 task is quite different from and much more challenging than previous tasks that use states as observations~\cite{xu2022prompting} or take into robot proprioception~\cite{r3m}. We only utilize single-frame visual input and instructions. The Flat R3M method, lacking language conditioning and skill sets can only succeed on tasks like \emph{drawer-close} and \emph{reach} through behavior cloning. Lang-conditioned Diffuser and modified LISA both outperform Flat R3M, suggesting the value of each corresponding module. 
% In contrast, 
Our \alias, discretizing skills into a skill set, achieves a 6\% higher performance than language-conditioned diffuser and a 9.5\% higher than LISA, demonstrating the effectiveness of this combinational architecture. 

\vspace{-1pt}
\subsection{Ablation Study on Reusability of Learned Skills}
To evaluate the reusability of our learned skills, we calculate the average number of different skills used for a single instruction and the total number of instructions using each skill of \lorl Sawyer Dataset in Table~\ref{tab:freq_lorel}. (With max episode step being 20, we experiment with skill horizon 10.)
We observe each single instruction uses 1.55 sub-skills on average and each skill is called multiple times than the number of instructions (75 with 5 eval episodes), verifying the transferability of learnt skills. As suggested in Tab.10 of~\cite{lisa}, except a very small horizon will hurt the performance, learning sub-skills to get refined semantics helps perform different actions at different stages. Besides, we also visualize resulting images from applying discrete skills in Appendix~\ref{append:result_image} to further validate skills' interpretability.

% \vspace{0.1cm}
% \noindent \textbf{Ablation on Inverse Dynamics and Loss Settings.}

% \begin{table}[tb]
%     \centering
% 	\small
% % 	\vskip-3pt
% % 	\tabcolsep 3pt
% 	% \vskip-5pt
% 	\begin{tabular}{lccc}
%  \toprule
% 		\textbf{Index} & \textbf{Inv setting} & \textbf{Loss Setting} & \textbf{Performance} \\ 
%   \midrule
% 		Setting 1 & w/o & \ding{55} & 18.0\% \\
% 		Setting 2 &  w/ obs feat  & \ding{55} & 16.7\% \\
% 		Setting 3  & w/ obs & \ding{55} & 20.0\% \\
% 		\alias & w/ obs  & \ding{51} & 23.3\% \\
%       % \midrule
%         % \multicolumn{1}{l}{\textbf{Average}} & 18.07 & 30.14 & \textbf{39.71} \hspace{.58cm} \\
% 		\bottomrule
%     \end{tabular}
%     \vspace{-5pt}
% 	\caption{\textbf{Ablation on Inverse Dynamics and Loss Settings.}}
%  \vspace{-10pt}
%  \label{tab:mateworld_res_2}
% \end{table}

\vspace{-1pt}
\subsection{Visualization Results of Learned Skill Set}
\begin{figure}[tb]
% \vspace{-2pt}
\centering 
\includegraphics[width=\linewidth]{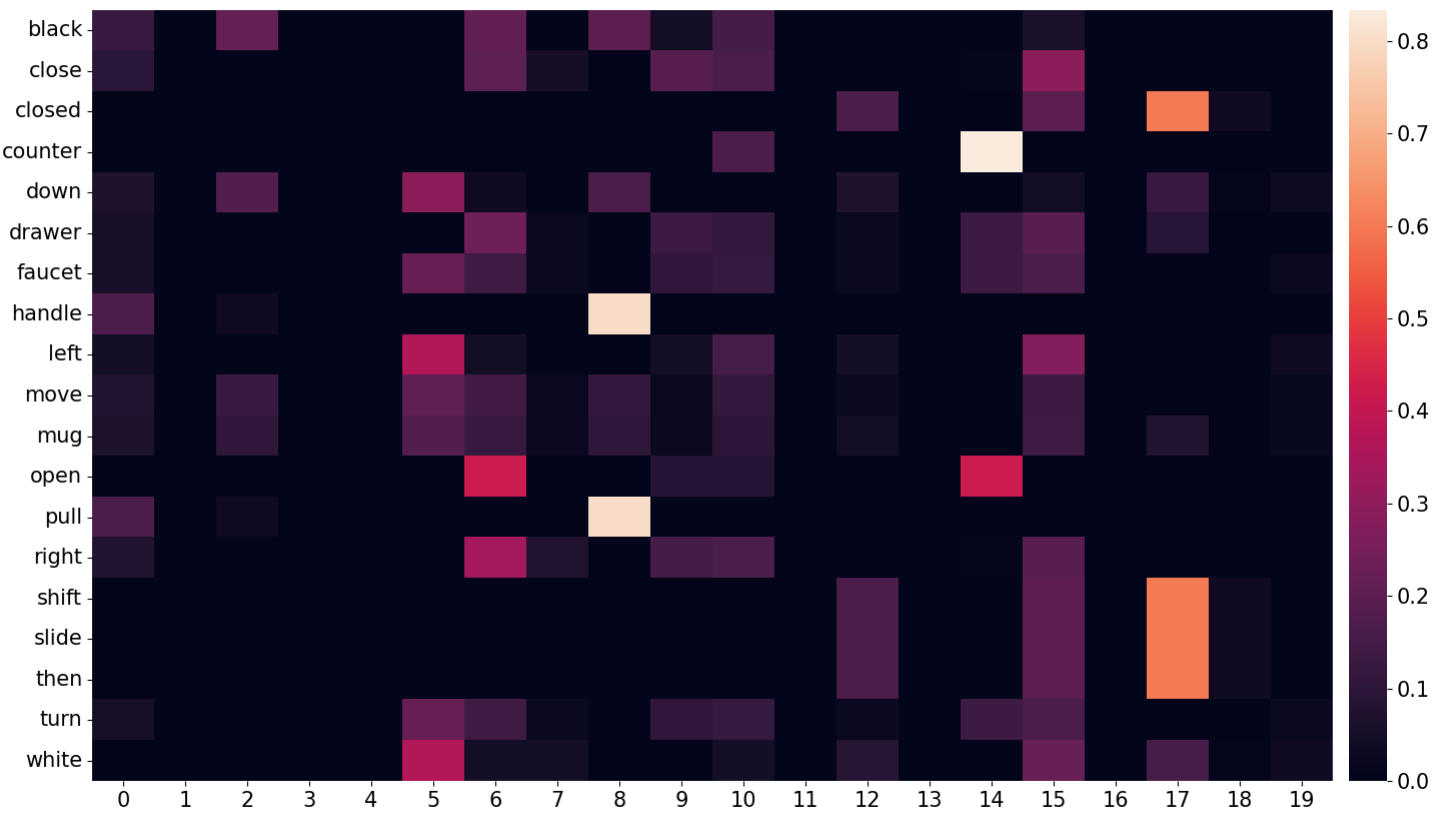}
\vspace{-23pt}
\caption{\textbf{Visualization of skill heat map on \lorl Sawyer compositional tasks.} We display the word frequency associated with a skill set of size 20 in \lorl, normalized by column. The data's sparsity and distinct highlights indicate certain language tokens are uniquely linked to specific skills. There are eleven skills learned by our method. (zoom in for best view)}
\vspace{-15pt}
\label{fig:heatmap_combi} 
\end{figure}

We show the visualization of skill set on \lorl compositional tasks here in Fig.~\ref{fig:heatmap_combi} and results on original \lorl dataset in Fig.~\ref{fig:heatmap_lorel} and \metaworld dataset in Fig.~\ref{fig:heatmap_MT10} in Appendix~\ref{append:visualize}.
The visual analysis of our \alias's learned skills on \lorl compositional tasks reveals that out of the 20-size skill set, our method learned 11 skills (\eg \emph{pull handle} [skill 0], \emph{open counter} [skill 14], \etc) notably distinguished by their unique word highlights. These bright spots across eleven columns (changing from only one column at initial which corresponds to default BC) in the heatmap underscore the model's ability to identify and isolate distinct skills from visual inputs, without an explicitly defined skill library. This indicates not only a significant interpretative advancement over previous diffusion-based planning but a successful abstraction of high-level skill representations. 
% Such clear skill differentiation demonstrates our method's efficacy in learning and encoding a diverse set of actions, paving the way for more intuitive human-AI interactions.

\begin{table}[tb]
\centering
\small
% \scriptsize
% \renewcommand\arraystretch{.1}
% \tabcolsep3.5pt
\resizebox{1.05\linewidth}{!}{
\begin{tabular}{c | c | c | c | c | c}
\toprule
\textbf{\# of learnt skills} & \textbf{\# of inst} & \textbf{\# of success} & \textbf{Use 1 skill} & \textbf{Use 2 skills} & \textbf{Average}\\
\midrule
17 & 375 & 144 & 64 & 80 & 1.55\\
\midrule
\textbf{Freq of 17 skills} & \multicolumn{5}{c}{30, 8, 14, 10, 5, 19, 4, 7, 20, 2, 9, 20, 25, 6, 10, 3, 18}\\
\bottomrule
\end{tabular}}
  \vspace{-8pt}
  \caption{\textbf{Average number of different skills used for a single instruction and total number of instructions used for each skill.}}
    \label{tab:freq_lorel}
    \vspace{-10pt}
\end{table}

\vspace{-5pt}
\section{Conclusion}
This paper presents \alias, an integrated framework that enables robots to perform tasks from natural language instructions by enabling interpretable skill learning and conditional diffusion planning. It employs vector quantization to learn discrete and comprehensible skill representations directly from visual and linguistic demonstrations. Subsequently, these skills condition a diffusion model to generate state trajectories adhering to the learned skills.
Through integrating hierarchical skill decomposition with conditional trajectory generation, \alias can comprehend and execute abstract instructions for various manipulation tasks.
% The modular design allows consistent latent planning across tasks by simply changing the inverse dynamics model. 
Extensive experiments on manipulation benchmarks demonstrate state-of-the-art performance, highlighting its effectiveness for multi-step composition tasks and ability to automatically learn interpretable skills.
% SkillDiffuser learns discrete yet human-understandable skill representations directly from visual and linguistic demonstrations using vector quantization. These skills then condition a diffusion model to generate customized state trajectories that adhere to the learned skills. By combining hierarchical skill decomposition with conditional trajectory generation in this manner, SkillDiffuser can follow abstract instructions to complete a variety of manipulation tasks. The modular design also allows for consistent latent planning across different tasks by simply changing the inverse dynamics model. Experiments on standard robotic manipulation benchmarks demonstrate state-of-the-art performance, highlighting its effectiveness for long-horizon tasks and its ability to automatically learn interpretable skills.
% The success of SkillDiffuser on challenging benchmarks and its ability to provide human-understandable skill representations paves the way for future developments in AI, where versatility, efficiency, and interpretability are paramount.
% Overall, SkillDiffuser represents a significant step towards intelligent agents that can acquire and deploy skills in an interpretable way.

\vspace{-5pt}
\section*{Acknowledgements}
This paper is partially supported by the National Key R\&D Program of China No.2022ZD0161000 and the General Research Fund of Hong Kong No.17200622.

\clearpage

% \clearpage
{
    \small\bibliographystyle{ieeenat_fullname}
    \bibliography{main}

\begin{thebibliography}{52}
\providecommand{\natexlab}[1]{#1}
\providecommand{\url}[1]{\texttt{#1}}
\expandafter\ifx\csname urlstyle\endcsname\relax
  \providecommand{\doi}[1]{doi: #1}\else
  \providecommand{\doi}{doi: \begingroup \urlstyle{rm}\Url}\fi

\bibitem[Ahn et~al.(2022)Ahn, Brohan, Brown, Chebotar, Cortes, David, Finn, Fu, Gopalakrishnan, Hausman, et~al.]{saycan}
Michael Ahn, Anthony Brohan, Noah Brown, Yevgen Chebotar, Omar Cortes, Byron David, Chelsea Finn, Chuyuan Fu, Keerthana Gopalakrishnan, Karol Hausman, et~al.
\newblock Do as i can, not as i say: Grounding language in robotic affordances.
\newblock 2022.

\bibitem[Ajay et~al.(2022)Ajay, Du, Gupta, Tenenbaum, Jaakkola, and Agrawal]{decisiondiffuser}
Anurag Ajay, Yilun Du, Abhi Gupta, Joshua~B Tenenbaum, Tommi~S Jaakkola, and Pulkit Agrawal.
\newblock Is conditional generative modeling all you need for decision making?
\newblock In \emph{The Eleventh International Conference on Learning Representations}, 2022.

\bibitem[Carvalho et~al.(2023)Carvalho, Le, Baierl, Koert, and Peters]{carvalho2023motion}
Joao Carvalho, An~T Le, Mark Baierl, Dorothea Koert, and Jan Peters.
\newblock Motion planning diffusion: Learning and planning of robot motions with diffusion models.
\newblock \emph{arXiv preprint arXiv:2308.01557}, 2023.

\bibitem[Chen et~al.(2021)Chen, Lu, Rajeswaran, Lee, Grover, Laskin, Abbeel, Srinivas, and Mordatch]{DT}
Lili Chen, Kevin Lu, Aravind Rajeswaran, Kimin Lee, Aditya Grover, Misha Laskin, Pieter Abbeel, Aravind Srinivas, and Igor Mordatch.
\newblock Decision transformer: Reinforcement learning via sequence modeling.
\newblock \emph{Advances in neural information processing systems}, 34:\penalty0 15084--15097, 2021.

\bibitem[Chi et~al.(2023)Chi, Feng, Du, Xu, Cousineau, Burchfiel, and Song]{chi2023diffusionpolicy}
Cheng Chi, Siyuan Feng, Yilun Du, Zhenjia Xu, Eric Cousineau, Benjamin Burchfiel, and Shuran Song.
\newblock Diffusion policy: Visuomotor policy learning via action diffusion.
\newblock In \emph{Proceedings of Robotics: Science and Systems (RSS)}, 2023.

\bibitem[Dhariwal and Nichol(2021)]{dhariwal2021diffusion}
Prafulla Dhariwal and Alexander Nichol.
\newblock Diffusion models beat gans on image synthesis.
\newblock \emph{Advances in neural information processing systems}, 34:\penalty0 8780--8794, 2021.

\bibitem[Du et~al.(2023)Du, Yang, Dai, Dai, Nachum, Tenenbaum, Schuurmans, and Abbeel]{du2023learning}
Yilun Du, Mengjiao Yang, Bo Dai, Hanjun Dai, Ofir Nachum, Joshua~B Tenenbaum, Dale Schuurmans, and Pieter Abbeel.
\newblock Learning universal policies via text-guided video generation.
\newblock \emph{Advances in Neural Information Processing Systems}, 2023.

\bibitem[Duan et~al.(2017)Duan, Andrychowicz, Stadie, Jonathan~Ho, Schneider, Sutskever, Abbeel, and Zaremba]{duan2017one}
Yan Duan, Marcin Andrychowicz, Bradly Stadie, OpenAI Jonathan~Ho, Jonas Schneider, Ilya Sutskever, Pieter Abbeel, and Wojciech Zaremba.
\newblock One-shot imitation learning.
\newblock \emph{Advances in neural information processing systems}, 30, 2017.

\bibitem[Eysenbach et~al.(2018)Eysenbach, Gupta, Ibarz, and Levine]{eysenbach2018diversity}
Benjamin Eysenbach, Abhishek Gupta, Julian Ibarz, and Sergey Levine.
\newblock Diversity is all you need: Learning skills without a reward function.
\newblock \emph{arXiv preprint arXiv:1802.06070}, 2018.

\bibitem[Feller(2015)]{feller2015theory}
William Feller.
\newblock On the theory of stochastic processes, with particular reference to applications.
\newblock In \emph{Selected Papers I}, pages 769--798. Springer, 2015.

\bibitem[Finn et~al.(2017)Finn, Yu, Zhang, Abbeel, and Levine]{finn2017one}
Chelsea Finn, Tianhe Yu, Tianhao Zhang, Pieter Abbeel, and Sergey Levine.
\newblock One-shot visual imitation learning via meta-learning.
\newblock In \emph{Conference on robot learning}, pages 357--368. PMLR, 2017.

\bibitem[Garg et~al.(2021)Garg, Chakraborty, Cundy, Song, and Ermon]{garg2021iq}
Divyansh Garg, Shuvam Chakraborty, Chris Cundy, Jiaming Song, and Stefano Ermon.
\newblock Iq-learn: Inverse soft-q learning for imitation.
\newblock \emph{Advances in Neural Information Processing Systems}, 34:\penalty0 4028--4039, 2021.

\bibitem[Garg et~al.(2022)Garg, Vaidyanath, Kim, Song, and Ermon]{lisa}
Divyansh Garg, Skanda Vaidyanath, Kuno Kim, Jiaming Song, and Stefano Ermon.
\newblock Lisa: Learning interpretable skill abstractions from language.
\newblock \emph{Advances in Neural Information Processing Systems}, 35:\penalty0 21711--21724, 2022.

\bibitem[He et~al.(2023)He, Bai, Xu, Yang, Zhang, Wang, Zhao, and Li]{he2023diffusion}
Haoran He, Chenjia Bai, Kang Xu, Zhuoran Yang, Weinan Zhang, Dong Wang, Bin Zhao, and Xuelong Li.
\newblock Diffusion model is an effective planner and data synthesizer for multi-task reinforcement learning.
\newblock \emph{Advances in neural information processing systems}, 2023.

\bibitem[Heidergott(2007)]{taylor}
Bernd Heidergott, editor.
\newblock \emph{Taylor Series Expansions}, pages 179--263.
\newblock Springer US, Boston, MA, 2007.

\bibitem[Ho and Ermon(2016)]{ho2016generative}
Jonathan Ho and Stefano Ermon.
\newblock Generative adversarial imitation learning.
\newblock \emph{Advances in neural information processing systems}, 29, 2016.

\bibitem[Ho and Salimans(2021)]{ho2021classifier}
Jonathan Ho and Tim Salimans.
\newblock Classifier-free diffusion guidance.
\newblock In \emph{NeurIPS 2021 Workshop on Deep Generative Models and Downstream Applications}, 2021.

\bibitem[Ho et~al.(2020)Ho, Jain, and Abbeel]{ho2020denoising}
Jonathan Ho, Ajay Jain, and Pieter Abbeel.
\newblock Denoising diffusion probabilistic models.
\newblock \emph{Advances in neural information processing systems}, 33:\penalty0 6840--6851, 2020.

\bibitem[Janner et~al.(2022)Janner, Du, Tenenbaum, and Levine]{janner2022planning}
Michael Janner, Yilun Du, Joshua Tenenbaum, and Sergey Levine.
\newblock Planning with diffusion for flexible behavior synthesis.
\newblock In \emph{International Conference on Machine Learning}, pages 9902--9915. PMLR, 2022.

\bibitem[Jiang et~al.(2019)Jiang, Gu, Murphy, and Finn]{LCRL}
Yiding Jiang, Shixiang~Shane Gu, Kevin~P Murphy, and Chelsea Finn.
\newblock Language as an abstraction for hierarchical deep reinforcement learning.
\newblock \emph{Advances in Neural Information Processing Systems}, 32, 2019.

\bibitem[Kingma and Ba(2015)]{kingma2014adam}
Diederik~P Kingma and Jimmy Ba.
\newblock Adam: A method for stochastic optimization.
\newblock In \emph{International Conference on Learning Representations}, 2015.

\bibitem[Li et~al.(2019)Li, Florensa, Clavera, and Abbeel]{li2019sub}
Alexander~C Li, Carlos Florensa, Ignasi Clavera, and Pieter Abbeel.
\newblock Sub-policy adaptation for hierarchical reinforcement learning.
\newblock \emph{arXiv preprint arXiv:1906.05862}, 2019.

\bibitem[Liang et~al.(2023)Liang, Mu, Ding, Ni, Tomizuka, and Luo]{liang2023adaptdiffuser}
Zhixuan Liang, Yao Mu, Mingyu Ding, Fei Ni, Masayoshi Tomizuka, and Ping Luo.
\newblock Adaptdiffuser: Diffusion models as adaptive self-evolving planners.
\newblock In \emph{International Conference on Machine Learning}, 2023.

\bibitem[MacQueen et~al.(1967)]{kmeans}
James MacQueen et~al.
\newblock Some methods for classification and analysis of multivariate observations.
\newblock In \emph{Proceedings of the fifth Berkeley symposium on mathematical statistics and probability}, pages 281--297. Oakland, CA, USA, 1967.

\bibitem[Melnychuk et~al.(2022)Melnychuk, Frauen, and Feuerriegel]{melnychuk2022causal}
Valentyn Melnychuk, Dennis Frauen, and Stefan Feuerriegel.
\newblock Causal transformer for estimating counterfactual outcomes.
\newblock In \emph{International Conference on Machine Learning}, pages 15293--15329. PMLR, 2022.

\bibitem[Misra(2020)]{misra2019mish}
Diganta Misra.
\newblock Mish: A self regularized non-monotonic activation function.
\newblock In \emph{31st British Machine Vision Conference 2020, {BMVC} 2020, Virtual Event, UK, September 7-10, 2020}, 2020.

\bibitem[Nachum et~al.(2018)Nachum, Gu, Lee, and Levine]{nachum2018data}
Ofir Nachum, Shixiang~Shane Gu, Honglak Lee, and Sergey Levine.
\newblock Data-efficient hierarchical reinforcement learning.
\newblock \emph{Advances in neural information processing systems}, 31, 2018.

\bibitem[Nair et~al.(2018)Nair, Pong, Dalal, Bahl, Lin, and Levine]{nair2018visual}
Ashvin~V Nair, Vitchyr Pong, Murtaza Dalal, Shikhar Bahl, Steven Lin, and Sergey Levine.
\newblock Visual reinforcement learning with imagined goals.
\newblock \emph{Advances in neural information processing systems}, 31, 2018.

\bibitem[Nair et~al.(2022)Nair, Mitchell, Chen, Savarese, Finn, et~al.]{lorl}
Suraj Nair, Eric Mitchell, Kevin Chen, Silvio Savarese, Chelsea Finn, et~al.
\newblock Learning language-conditioned robot behavior from offline data and crowd-sourced annotation.
\newblock In \emph{Conference on Robot Learning}, pages 1303--1315. PMLR, 2022.

\bibitem[Nair et~al.(2023)Nair, Rajeswaran, Kumar, Finn, and Gupta]{r3m}
Suraj Nair, Aravind Rajeswaran, Vikash Kumar, Chelsea Finn, and Abhinav Gupta.
\newblock R3m: A universal visual representation for robot manipulation.
\newblock In \emph{Conference on Robot Learning}, pages 892--909. PMLR, 2023.

\bibitem[Ni et~al.(2023)Ni, Hao, Mu, Yuan, Zheng, Wang, and Liang]{ni2023metadiffuser}
Fei Ni, Jianye Hao, Yao Mu, Yifu Yuan, Yan Zheng, Bin Wang, and Zhixuan Liang.
\newblock Metadiffuser: Diffusion model as conditional planner for offline meta-rl.
\newblock In \emph{International Conference on Machine Learning}, 2023.

\bibitem[Niekum et~al.(2015)Niekum, Osentoski, Konidaris, Chitta, Marthi, and Barto]{niekum2015learning}
Scott Niekum, Sarah Osentoski, George Konidaris, Sachin Chitta, Bhaskara Marthi, and Andrew~G Barto.
\newblock Learning grounded finite-state representations from unstructured demonstrations.
\newblock \emph{The International Journal of Robotics Research}, 34\penalty0 (2):\penalty0 131--157, 2015.

\bibitem[Pomerleau(1991)]{pomerleau1991bc}
Dean~A Pomerleau.
\newblock Efficient training of artificial neural networks for autonomous navigation.
\newblock \emph{Neural computation}, 3\penalty0 (1):\penalty0 88--97, 1991.

\bibitem[Puterman(1994)]{puterman1994markov}
Martin~L Puterman.
\newblock \emph{Markov decision processes: discrete stochastic dynamic programming}.
\newblock John Wiley \& Sons, 1994.

\bibitem[Radford et~al.(2019)Radford, Wu, Child, Luan, Amodei, Sutskever, et~al.]{gpt2}
Alec Radford, Jeffrey Wu, Rewon Child, David Luan, Dario Amodei, Ilya Sutskever, et~al.
\newblock Language models are unsupervised multitask learners.
\newblock \emph{OpenAI blog}, 1\penalty0 (8):\penalty0 9, 2019.

\bibitem[Ronneberger et~al.(2015)Ronneberger, Fischer, and Brox]{ronneberger2015u}
Olaf Ronneberger, Philipp Fischer, and Thomas Brox.
\newblock U-net: Convolutional networks for biomedical image segmentation.
\newblock In \emph{Medical Image Computing and Computer-Assisted Intervention--MICCAI 2015: 18th International Conference, Munich, Germany, October 5-9, 2015, Proceedings, Part III 18}, pages 234--241. Springer, 2015.

\bibitem[Ross and Bagnell(2010)]{ross2010efficient}
St{\'e}phane Ross and Drew Bagnell.
\newblock Efficient reductions for imitation learning.
\newblock In \emph{Proceedings of the thirteenth international conference on artificial intelligence and statistics}, pages 661--668. JMLR Workshop and Conference Proceedings, 2010.

\bibitem[Ross et~al.(2011)Ross, Gordon, and Bagnell]{ross2011reduction}
St{\'e}phane Ross, Geoffrey Gordon, and Drew Bagnell.
\newblock A reduction of imitation learning and structured prediction to no-regret online learning.
\newblock In \emph{Proceedings of the fourteenth international conference on artificial intelligence and statistics}, pages 627--635. JMLR Workshop and Conference Proceedings, 2011.

\bibitem[Sanh(2019)]{distilbert}
V Sanh.
\newblock Distilbert, a distilled version of bert: smaller, faster, cheaper and lighter.
\newblock In \emph{Proceedings of Thirty-third Conference on Neural Information Processing Systems}, 2019.

\bibitem[Sharma et~al.(2019)Sharma, Gu, Levine, Kumar, and Hausman]{sharma2019dynamics}
Archit Sharma, Shixiang Gu, Sergey Levine, Vikash Kumar, and Karol Hausman.
\newblock Dynamics-aware unsupervised discovery of skills.
\newblock \emph{arXiv preprint arXiv:1907.01657}, 2019.

\bibitem[Singh et~al.(2020)Singh, Jang, Irpan, Kappler, Dalal, Levinev, Khansari, and Finn]{singh2020scalable}
Avi Singh, Eric Jang, Alexander Irpan, Daniel Kappler, Murtaza Dalal, Sergey Levinev, Mohi Khansari, and Chelsea Finn.
\newblock Scalable multi-task imitation learning with autonomous improvement.
\newblock In \emph{2020 IEEE International Conference on Robotics and Automation (ICRA)}, pages 2167--2173. IEEE, 2020.

\bibitem[Stepputtis et~al.(2020)Stepputtis, Campbell, Phielipp, Lee, Baral, and Ben~Amor]{LCBC}
Simon Stepputtis, Joseph Campbell, Mariano Phielipp, Stefan Lee, Chitta Baral, and Heni Ben~Amor.
\newblock Language-conditioned imitation learning for robot manipulation tasks.
\newblock \emph{Advances in Neural Information Processing Systems}, 33:\penalty0 13139--13150, 2020.

\bibitem[Stuart and Ord(1994)]{bayes}
Alan Stuart and J~Keith Ord.
\newblock Kendall's advanced theory of statistics. vol. 1: Distribution theory.
\newblock \emph{Kendall's advanced theory of statistics. Vol. 1: Distribution theory}, 1994.

\bibitem[Torabi et~al.(2018)Torabi, Warnell, and Stone]{torabi2018behavioral}
Faraz Torabi, Garrett Warnell, and Peter Stone.
\newblock Behavioral cloning from observation.
\newblock In \emph{Proceedings of the 27th International Joint Conference on Artificial Intelligence}, pages 4950--4957, 2018.

\bibitem[Van Den~Oord et~al.(2017)Van Den~Oord, Vinyals, et~al.]{vq-vae}
Aaron Van Den~Oord, Oriol Vinyals, et~al.
\newblock Neural discrete representation learning.
\newblock \emph{Advances in neural information processing systems}, 30, 2017.

\bibitem[Vaswani et~al.(2017)Vaswani, Shazeer, Parmar, Uszkoreit, Jones, Gomez, Kaiser, and Polosukhin]{transformer}
Ashish Vaswani, Noam Shazeer, Niki Parmar, Jakob Uszkoreit, Llion Jones, Aidan~N Gomez, {\L}ukasz Kaiser, and Illia Polosukhin.
\newblock Attention is all you need.
\newblock \emph{Advances in neural information processing systems}, 30, 2017.

\bibitem[Wu and He(2018)]{wu2018group}
Yuxin Wu and Kaiming He.
\newblock Group normalization.
\newblock In \emph{Proceedings of the European conference on computer vision (ECCV)}, pages 3--19, 2018.

\bibitem[Xu et~al.(2018)Xu, Nair, Zhu, Gao, Garg, Fei-Fei, and Savarese]{xu2018neural}
Danfei Xu, Suraj Nair, Yuke Zhu, Julian Gao, Animesh Garg, Li Fei-Fei, and Silvio Savarese.
\newblock Neural task programming: Learning to generalize across hierarchical tasks.
\newblock In \emph{2018 IEEE International Conference on Robotics and Automation (ICRA)}, pages 3795--3802. IEEE, 2018.

\bibitem[Xu et~al.(2022)Xu, Shen, Zhang, Lu, Zhao, Tenenbaum, and Gan]{xu2022prompting}
Mengdi Xu, Yikang Shen, Shun Zhang, Yuchen Lu, Ding Zhao, Joshua Tenenbaum, and Chuang Gan.
\newblock Prompting decision transformer for few-shot policy generalization.
\newblock In \emph{international conference on machine learning}, pages 24631--24645. PMLR, 2022.

\bibitem[Yang et~al.(2023)Yang, Huang, Lei, Zhong, Yang, Fang, Wen, Zhou, and Lin]{yang2023policy}
Long Yang, Zhixiong Huang, Fenghao Lei, Yucun Zhong, Yiming Yang, Cong Fang, Shiting Wen, Binbin Zhou, and Zhouchen Lin.
\newblock Policy representation via diffusion probability model for reinforcement learning.
\newblock \emph{arXiv preprint arXiv:2305.13122}, 2023.

\bibitem[Yu et~al.(2020)Yu, Quillen, He, Julian, Hausman, Finn, and Levine]{yu2020meta}
Tianhe Yu, Deirdre Quillen, Zhanpeng He, Ryan Julian, Karol Hausman, Chelsea Finn, and Sergey Levine.
\newblock Meta-world: A benchmark and evaluation for multi-task and meta reinforcement learning.
\newblock In \emph{Conference on robot learning}, pages 1094--1100. PMLR, 2020.

\bibitem[Zhang et~al.(2021)Zhang, Yu, and Xu]{zhang2021hierarchical}
Jesse Zhang, Haonan Yu, and Wei Xu.
\newblock Hierarchical reinforcement learning by discovering intrinsic options.
\newblock \emph{arXiv preprint arXiv:2101.06521}, 2021.

\end{thebibliography}
}

\clearpage
\setcounter{page}{1}
\maketitlesupplementary

\appendix

\section{Theoretical Foundation of Classifier-free Diffusion Model for Planning}
\label{append:math}
\subsection{Review of Classifier-guided Diffusion Model}
Firstly, for a given trajectory $\btau$, the standard reverse process of an unconditional diffusion probabilistic model is defined by $p_{\theta}(\btau^i|\btau^{i+1})$. This framework is then extended to incorporate conditioning on a specific label $\by$ (\eg, the reward), which is considered in the context of current-step denoised trajectory \(\btau^i\), which is represented as \(p_{\phi}(\by|\btau^i)\). Consequently, the reverse diffusion process can be reformulated as \(p_{\theta,\phi}(\btau^i|\btau^{i+1},\by)\). This approach introduces additional parameters \(\phi\) alongside the original diffusion model parameters \(\theta\). The parameters \(\phi\) can be viewed as a classifier, that encapsulates the probability of whether a noisy trajectory \(\btau^i\) satisfies the specific label \(\by\), with a notation of \(p_{\phi}(\by|\btau^i)\).

Under the constraints illustrated in~\cite{liang2023adaptdiffuser,dhariwal2021diffusion}, we can derive the following theorem with lemma 
\begin{equation}
    p_{\theta, \phi}\left(\by \mid \btau^{i}, \btau^{i+1}\right) = p_{\phi}\left(\by \mid \btau^{i}\right).
\end{equation}

\begin{theorem}
The conditional sampling probability of reverse diffusion process $p_{\theta,\phi}(\btau^i \mid \btau^{i+1},\by)$ is proportional to unconditional transition probability $p_{\theta}(\btau^i \mid \btau^{i+1})$ multiplied by the classified probability $p_{\phi}(\by \mid \btau^i)$.
\begin{equation}
    p_{\theta,\phi}(\btau^i \mid \btau^{i+1},\by) = Z p_{\theta}(\btau^i \mid \btau^{i+1})p_{\phi}(\by \mid \btau^i)
\label{eq:theorem1}
\end{equation}
\end{theorem}

\begin{proof}
% With the knowledge of conditional probability and Bayes' theorem, we have,
\begin{equation}
    \begin{split}
p_{\theta,\phi}(\btau^i \mid \btau^{i+1},&\ \by) = \frac{p_{\theta, \phi}\left(\btau^{i}, \btau^{i+1}, \by\right)}{p_{\theta, \phi}\left(\btau^{i+1}, \by\right)} 
\\
% & =\frac{p_{\theta, \phi}\left(\btau^{i}, \btau^{i+1}, \by\right)}{p_{\phi}\left(\by \mid \btau^{i+1}\right) p_{\theta}\left(\btau^{i+1}\right)} 
% \\
& =\frac{p_{\theta, \phi}\left(\by \mid \btau^{i}, \btau^{i+1}\right) p_{\theta}\left(\btau^{i}, \btau^{i+1}\right)}{p_{\phi}\left(\by \mid \btau^{i+1}\right) p_{\theta}\left(\btau^{i+1}\right)} 
\\
& =\frac{p_{\theta, \phi}\left(\by \mid \btau^{i}, \btau^{i+1}\right) p_{\theta}\left(\btau^{i} \mid \btau^{i+1}\right) p_{\theta}\left(\btau^{i+1}\right)}{p_{\phi}\left(\by \mid \btau^{i+1}\right) p_{\theta}\left(\btau^{i+1}\right)} 
\\
& =\frac{p_{\theta, \phi}\left(\by \mid \btau^{i}, \btau^{i+1}\right) p_{\theta}\left(\btau^{i} \mid \btau^{i+1}\right)}{p_{\phi}\left(\by \mid \btau^{i+1}\right)} 
\\
& =\frac{p_{\phi}\left(\by \mid \btau^{i}\right) p_{\theta}\left(\btau^{i} \mid \btau^{i+1}\right)}{p_{\phi}\left(\by \mid \btau^{i+1}\right)}. 
\end{split}
\label{eq:append-guided-diff}
\end{equation}

The term $p_{\phi}\left(\by \mid \btau^{i+1}\right)$ is not directly correlated to $\btau^{i}$ at the diffusion timestep $i$, thus can be viewed as a constant with notation $Z$.
\end{proof}

On this basis, using Taylor series expansion~\cite{taylor}, we can sample trajectories by the modified Gaussian resampling.
\begin{theorem}
With a sufficiently large number of reverse diffusion steps, the sampling from reverse diffusion process $p_{\theta,\phi}(\btau^i \mid \btau^{i+1},\by)$ can be approximated by a modified Gaussian resampling. That is
\begin{equation}
    p_{\theta,\phi}(\btau^i|\btau^{i+1},\by) \approx \mathcal{N}(\btau^i; \mu_{\theta} + \Sigma \nabla_{\btau} \log p_{\phi}\left(\by \mid \btau^{i}\right), \Sigma),
    \label{eq:theorem2}
\end{equation}
where $\mu_{\theta}=\mu_{\theta}(\btau^i)$ and $\Sigma$ are the mean and variance of unconditional reverse diffusion process $p_{\theta}(\btau^i \mid \btau^{i+1})$.
\end{theorem}

\begin{proof}
With the above definition, we can rewrite the transfer probability of the unconditional denoising process as 
\begin{align}
    p_{\theta}(\btau^i \mid \btau^{i+1}) &= \mathcal{N}(\btau^i; \mu_{\theta}, \Sigma) \\
    \log p_{\theta}(\btau^i \mid \btau^{i+1}) &= -\frac{1}{2}(\btau^i - \mu_{\theta})^T \Sigma^{-1} (\btau^{i} - \mu_\theta) + C
    \label{eq:tt1}
\end{align}

With a sufficiently large number of reverse diffusion steps, we apply Taylor expansion around $\btau^i=\mu_{\theta}$ as
\begin{align*}
    \log p_{\phi}\left(\by \mid \btau^{i}\right) &= \log p_{\phi}\left(\by \mid \btau^{i}\right)|_{\btau^{i}=\mu_{\theta}}\\
    &\hspace{10pt}+\left.\left(\btau^{i}-\mu_{\theta}\right) \nabla_{\btau^{i}} \log p_{\phi}\left(\by \mid \btau^{i}\right)\right|_{\btau^{i}=\mu_{\theta}}.
    % \label{eq:tt2}
\end{align*}
Therefore, using Eq.~\ref{eq:append-guided-diff}, we derive
\begin{equation*}
    \log p_{\theta,\phi}(\btau^i |\btau^{i+1},\by) = \log p_{\theta}(\btau^i|\btau^{i+1}) + \log p_{\phi}(\by|\btau^i)+C_1
\end{equation*}
\begin{equation}
\begin{split}
    RHS & = -\frac{1}{2}\left(\btau^{i}-\mu_{\theta}\right)^{T} \Sigma^{-1}\left(\btau^{i}-\mu_{\theta}\right)\\
         &\hspace{10pt}+\left(\btau^{i}-\mu_{\theta}\right) \nabla \log p_{\phi}\left(\by \mid \btau^{i}\right) +C_{2}
    \end{split}
\end{equation}
\begin{equation*}
\begin{split}
    RHS & =-\frac{1}{2}\left(\btau^{i}-\mu_{\theta}-\Sigma \nabla \log p_{\phi}\left(\by \mid \btau^{i}\right)\right)^{T} \times\Sigma^{-1}\\
    &\hspace{10pt}\times\left(\btau^{i}-\mu_{\theta}-\Sigma \nabla \log p_{\phi}\left(\by \mid \btau^{i}\right)\right)+C_{3},
\end{split}
\end{equation*}
which means,
\begin{equation*}
    p_{\theta,\phi}(\btau^i|\btau^{i+1},\by) \approx \mathcal{N}(\btau^i; \mu_{\theta} + \Sigma \nabla_{\btau} \log p_{\phi}\left(\by \mid \btau^{i}\right), \Sigma)
\end{equation*}
\end{proof}

\subsection{Classifier-free Diffusion Model}
While classifier guidance successfully achieves conditional guidance during trajectory generation, it is nonetheless reliant on gradients from a separate trained classifier which is hard to obtain in many cases. Classifier-free guidance~\cite{ho2020denoising} seeks to eliminate the classifier, which achieves the same effect as classifier guidance, but without such gradients. 

First of all, we define the score function of the unconditional diffusion model as 
\begin{equation}
    \epsilon_{\theta}(\btau^i)=-\Sigma \nabla_{\btau} \log p_{\phi}\left(\btau^{i}\right).
    \label{eq:append-16}
\end{equation}

Then, through Eq.~\ref{eq:theorem1} and~\ref{eq:theorem2}, the score function of the classifier-guided diffusion model can be expressed as
\begin{equation}
    \epsilon_{\theta}(\btau^i, \by)=\epsilon_{\theta}(\btau^i) - \alpha\Sigma \nabla_{\btau} \log p_{\phi}\left(\by \mid \btau^{i}\right),
    \label{eq:append-17}
\end{equation}
where $\alpha$ is a scale hyper-parameter.

\begin{theorem}
Classifier-free guided diffusion model performs sampling with the linear combination of the conditional and unconditional score estimates as,
\begin{equation}
\hat{\epsilon}_{\theta} = \hat{\epsilon}_{\theta}(\btau^i, \by)= (1-\omega)\epsilon_\theta(\btau^i) + \omega\epsilon_\theta(\btau^i, \boldsymbol{y}),
\label{eq:append-18}
\end{equation}
implicitly embedding guidance into the score function, with $\omega$ the scale hyper-parameter.
\end{theorem}
\begin{proof}
    Considering there is an implicit classifier denoted as $\tilde{p}_{\phi}(\by\mid \btau^i)$, with Bayes Rule~\cite{bayes}, we can expand it as
    \begin{equation*}
        \tilde{p}_{\phi}(\by\mid \btau^i) \propto \tilde{p}_{\theta,\phi}(\btau^i, \by) / p_{\theta}(\btau^i).
    \end{equation*}
    
    Then gradient of this implicit classifier would be
    \begin{equation}
        \nabla_{\btau}\log\tilde{p}_{\phi}(\by\mid \btau^i) = \nabla_{\btau}\log\tilde{p}_{\theta,\phi}(\btau^i, \by) - \nabla_{\btau}\log p_{\theta}(\btau^i).
    \end{equation}
    
    Substitute Eq.~\ref{eq:append-16} in RHS, we get
    \begin{align*}
        \alpha\Sigma\nabla_{\btau}\log\tilde{p}_{\phi}(\by\mid \btau^i) &= \alpha\Sigma\nabla_{\btau}\log\tilde{p}_{\theta,\phi}(\btau^i, \by)\\ 
        &\hspace{60pt}- \alpha\Sigma\nabla_{\btau}\log p_{\theta}(\btau^i)\\
        &= -\alpha\hat{\epsilon}_{\theta}(\btau^i, \by) + \alpha\epsilon_{\theta}(\btau^i)
    \end{align*}

    And then substitute Eq.~\ref{eq:append-17} in LHS,
    \begin{equation}
        \begin{split}
            \epsilon_{\theta}(\btau^i) - \epsilon_{\theta}(\btau^i, \by) &= -\alpha\hat{\epsilon}_{\theta}(\btau^i, \by) + \alpha\epsilon_{\theta}(\btau^i)\\
            \alpha\hat{\epsilon}_{\theta}(\btau^i, \by) &= \epsilon_{\theta}(\btau^i, \by) + (\alpha-1)\epsilon_{\theta}(\btau^i)\\
            \hat{\epsilon}_{\theta}(\btau^i, \by) = (1/&\alpha)\epsilon_{\theta}(\btau^i, \by) + (1-1/\alpha)\epsilon_{\theta}(\btau^i)
        \end{split}
    \end{equation}
    
    Let $\omega=1/\alpha$, we obtain,
    \begin{equation*}
        \hat{\epsilon}_{\theta}(\btau^i, \by) = (1-\omega)\epsilon_{\theta}(\btau^i) + \omega\epsilon_{\theta}(\btau^i, \by),
    \end{equation*}
    which is equal to Eq.~\ref{eq:append-18}.
\end{proof}
Therefore, in classifier-free diffusion guidance, we only need to train a single neural network to parameterize both conditional score estimator $\epsilon_{\theta}(\btau^i, \by)$ and unconditional score estimator $\epsilon_{\theta}(\btau^i)$, where for the unconditional model we can set an empty set $\varnothing$ for the condition identifier $\by$ when predicting the score, \ie $\epsilon_{\theta}(\btau^i) = \epsilon_{\theta}(\btau^i, \by=\varnothing)$. Following the settings of~\cite{ho2020denoising}, we jointly train the unconditional and conditional models simply by randomly setting $\by$ to the unconditional class identifier $\varnothing$ with probability $\beta$, which balances off the diversity and the relevance of the conditional label of generated samples.

% Classifier-free guidance is an alternative
% method of modifying θ(zλ, c) to have the same effect as classifier guidance, but without a classifier.

\section{Pseudo-code of Training \alias}
\label{append:pseudo}
As illustrated in Sec.~\ref{sec:train_process}, we provide the pseudocode for our \alias's training process in Algorithm \ref{alg:training}, detailing its sequential stages and core mechanics. Additionally, Algorithm \ref{alg:inference} describes the inference process, illustrating its steps of skill abstraction and trajectory generation.

\begin{algorithm}
   \caption{Training process of \alias}
   \label{alg:training}
\begin{algorithmic}[1]
   \Require Dataset $\mathcal{D}$ of partially observed trajectories with paired language $\left\{\btau_{\xi} = (l, \{\vct{i}_t, \vct{a}_t\}_{t=0}^{T-1})\right\}_{\xi=1}^{N}$, size of the skill set $K$ and horizon $H$, pre-trained language and visual encoder $\Phi_{lang}$, $\Phi_{im}$
   \State Initialize skill predictor $f$, conditional diffusion model $\mathcal{M}$, skill embedding model $\Lambda$ and inverse dynamics model $\Psi$
   \State Vector Quantization op $\mathbf{q(\cdot)}$
  \While {\textit{not converged}}

   \State{Sample $\tau = (l, \{\vct{i}_t, \vct{a}_t\}_{t=0}^{T-1}$)}
   \State Initialize partially observed states $S=\{\Phi(\vct{i}_0)\}$
   \For{$k=0...\lfloor{\frac{T}{H}}\rfloor$} \Comment{Sample a skill every H steps}

%   \State $current\_traj = [s_0]$ \Comment{Initialize with start state}
%   \State  Append $s_0$ to $S$ \Comment{Add initial state}
%   \State $action\_preds = []$
%   \State $k=0$
%   \Repeat 
   \State  $z \leftarrow \mathbf{q}(f(\Phi_{lang}(l), S))$
   \State $\mathcal{L}_{\mathit{diff}} \leftarrow \mathcal{M}_{\mathit{diff}}(S, \Lambda(z))$ \Comment{Diffusing process}
   \For{\textit{step} $t=1...H$}
   % \Comment{Predict actions using a fixed skill and context length $H$}
   % \Comment{Use c $H$ steps}
   \State  $S \leftarrow S \cup \{\Phi(\vct{i}_{k H+t+1})\}$
   \State $\tilde{\ba}_{k H+t} \leftarrow \Psi([{\bs}_{k H+t}, {\bs}_{k H+t+1}], \vct{i}_{k H+t})$
   \Comment{Predict action using inverse dynamics model}
 % \Comment{Append seen state}
    \State $\mathcal{L}_{inv} = \mathbb{E}\left[\|\ba_{k H+t} - \tilde{\ba}_{k H+t}\|_2^2\right]$
    \State Train $\Psi$ with objective $\mathcal{L}_{inv}$
%   \State $action\_preds.append(a)$
%   \State $current\_traj.append(A[k*H+i])$
%   \State $current\_traj.append(S[k*H+i+1])$
%   \State $states.append(S[k*H+i+1])$
   \EndFor
%   \State $k+=1$
%   \Until{end of trajectory}
   \State Train $f, \Lambda$ and $\mathcal{M}$ with objective $\mathcal{L}_{\mathrm{VQ}} + \lambda \mathcal{L}_{\mathit{diff}}$
   \EndFor
   \EndWhile
\end{algorithmic}
\end{algorithm}
\vspace{-5pt}

\begin{algorithm}[h]
   \caption{Inference process of \alias}
   \label{alg:inference}
\begin{algorithmic}[1]
   \Require Initial partial observation $\vct{i}_0$ and the language instruction $l$, pre-trained language and visual encoder $\Phi_{lang}$, $\Phi_{im}$
   \Require Trained skill predictor $f$, conditional diffusion model $\mathcal{M}$, skill embedding model $\Lambda$ and inverse dynamics model $\Psi$

   \State Initialize partially observed states $S=\{\Phi(\vct{i}_0)\}$
   \For{$k=0...\lfloor{\frac{T}{H}}\rfloor$} \Comment{Sample a skill every H steps}

%   \State $current\_traj = [s_0]$ \Comment{Initialize with start state}
%   \State  Append $s_0$ to $S$ \Comment{Add initial state}
%   \State $action\_preds = []$
%   \State $k=0$
%   \Repeat 
   \State  $z \leftarrow \mathbf{q}(f(\Phi_{lang}(l), S))$
   \State $S' \leftarrow \mathcal{M}_{\mathit{denoise}}(S, \Lambda(z))$ \Comment{Denoising process}
   \For{\textit{step} $t=1...H$}
   % \Comment{Predict actions using a fixed skill and context length $H$}
   % \Comment{Use c $H$ steps}
   \State $\ba_{k H+t} \leftarrow \Psi([\bs_{k H+t}, \bs'_{k H+t+1}], \vct{i}_{k H+t})$
   % \Comment{Predict action using inverse dynamics model}
   \State $\tilde{\bs}_{k H+t+1} \leftarrow \text{Env.step}(\ba_{k H+t})$
   \Comment{Take action}
 % \Comment{Append seen state}
 \State  $S \leftarrow S \cup \{\tilde{\bs}_{k H+t+1}\}$
%   \State $action\_preds.append(a)$
%   \State $current\_traj.append(A[k*H+i])$
%   \State $current\_traj.append(S[k*H+i+1])$
%   \State $states.append(S[k*H+i+1])$
   \EndFor
%   \State $k+=1$
%   \Until{end of trajectory}
   \EndFor
\end{algorithmic}
\end{algorithm}

\begin{figure*}[htb]
\centering 
\includegraphics[width=\linewidth]{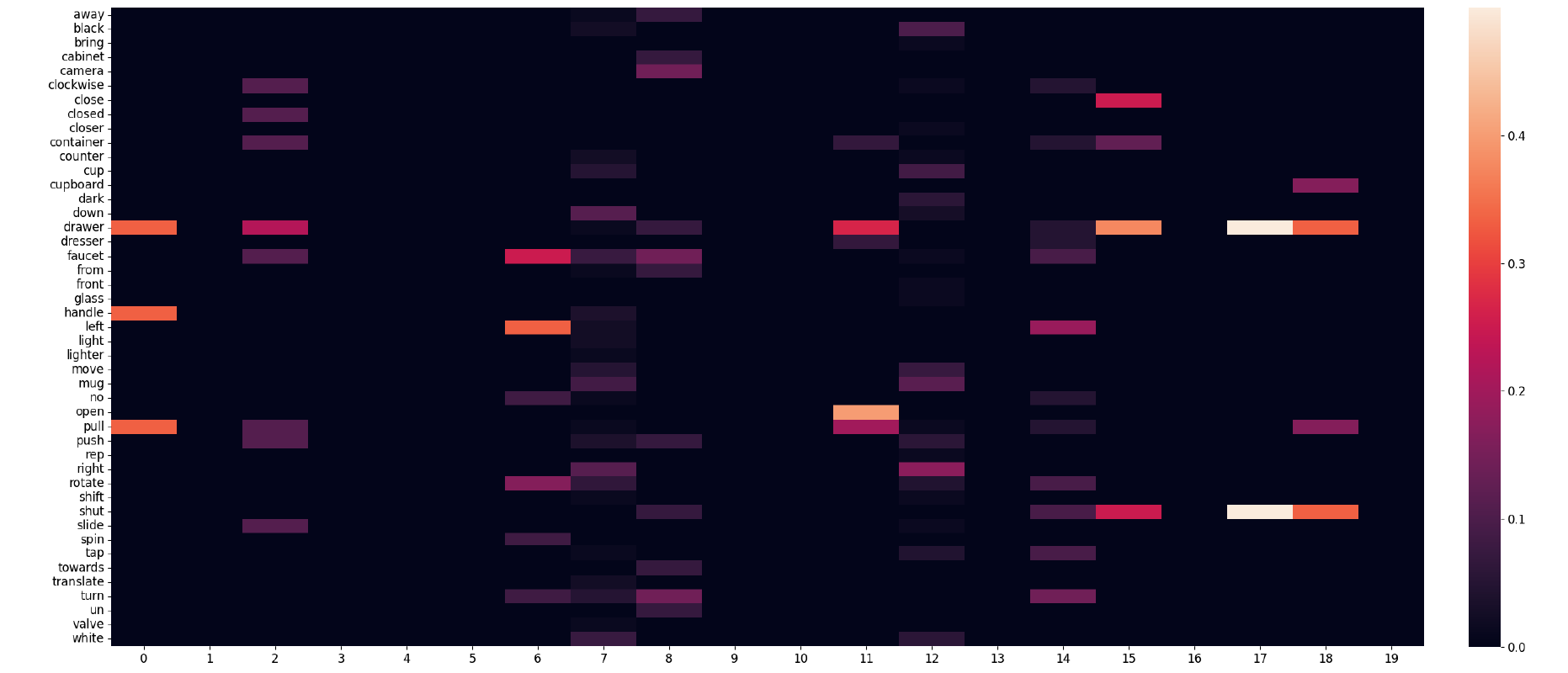}
\vspace{-20pt}
\caption{\textbf{Visualization of skill heat map on \lorl.} We display the word frequency associated with a skill set of size 20 in \lorl, normalized by column. The data's sparsity and distinct highlights indicate certain language tokens are uniquely linked to specific skills. There are eleven skills learned by our method.}
\vspace{-10pt}
\label{fig:heatmap_lorel} 
\end{figure*}

\begin{figure}[tb]
% \vspace{-2pt}
\centering 
\includegraphics[width=\linewidth]{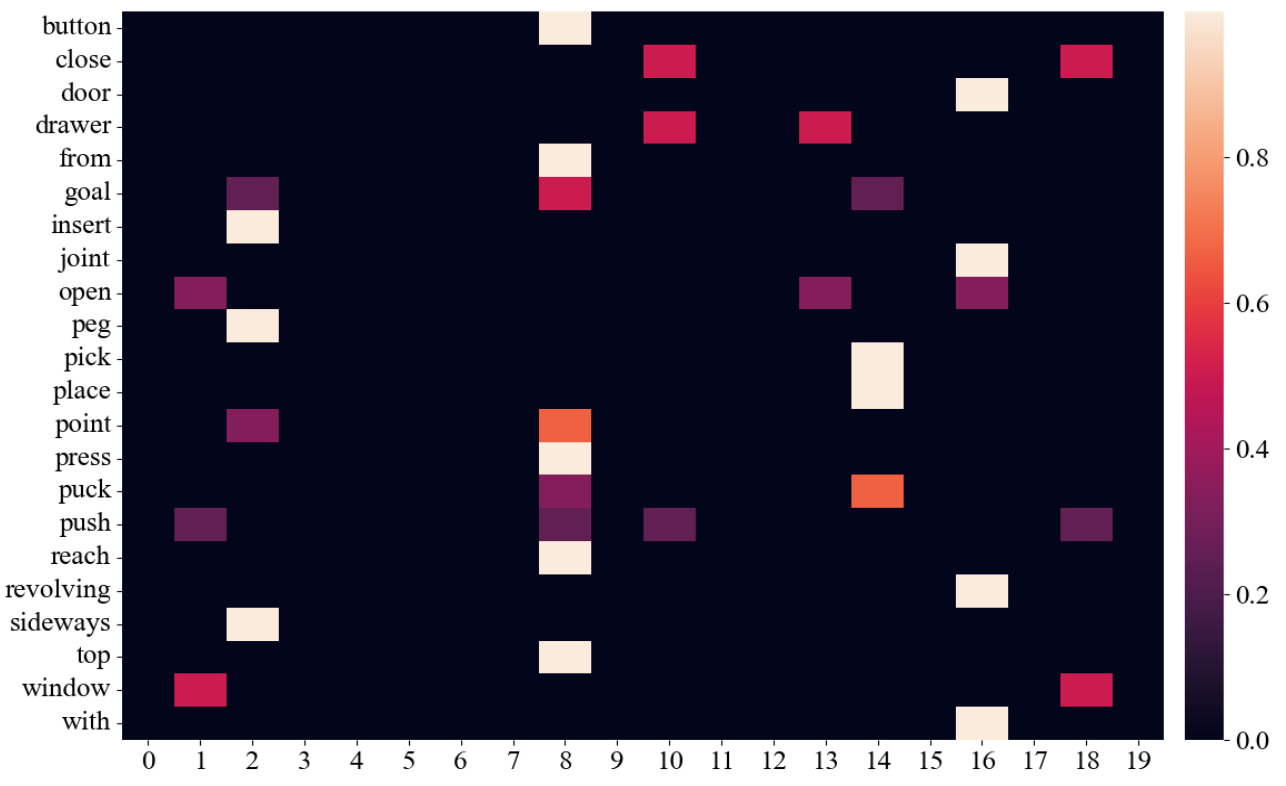}
\vspace{-20pt}
\caption{\textbf{Visualization of skill heat map on \metaworld Multi-Task 10 (MT10).} There are eight skills learned by our method. (zoom in for best view)}
\vspace{-15pt}
\label{fig:heatmap_MT10} 
\end{figure}

\section{More Visualizations}
\label{append:visualize}
\subsection{Visualization Results of Learned Skill Set}
% \textbf{Heat Map of Word Frequency in \lorl Dataset.}
As mentioned before, we show the visualization results of skill set on \lorl Sawyer Dataset in Fig.~\ref{fig:heatmap_lorel} and \metaworld Multi-Task 10 (MT10) in Fig.~\ref{fig:heatmap_MT10}. The visualization results show that out of a 20-size skill-set, our \alias learned 11 skills for \lorl (\eg \emph{pull drawer handle} [skill 0], \emph{shut close container drawer} [skill 15], \etc) and 8 skills for \metaworld MT10 (\eg \emph{open push window} [skill 0], \emph{open door with revolving joint} [skill 16], \etc). The results demonstrate strong skill abstraction abilities. For example, the skill ``shut close container drawer'' abstracts different expressions like ``shut drawer'', ``shut container'' into one skill semantic. In the heatmap, the presence of distinct bright spots across eleven columns strongly reaffirms the model's capability to discern and pinpoint specific skills from visual inputs, in the absence of a pre-defined skill library. This observation is not just a testament to the model's enhanced interpretative prowess over conventional diffusion-based planning approaches but also marks a remarkable stride in abstracting high-level skills into representations that are intuitively understandable by humans. Such evidence further validates the model's proficiency in sophisticated skill identification and representation.

% \vspace{2pt}
% \noindent \textbf{Heat Map of Skill Frequency in \lorl Sawyer Dataset.} We further show the heat map of skill frequency

\subsection{Word Cloud of Learned Skills}
\begin{figure*}[tb]
\centering 
\includegraphics[width=0.85\linewidth]{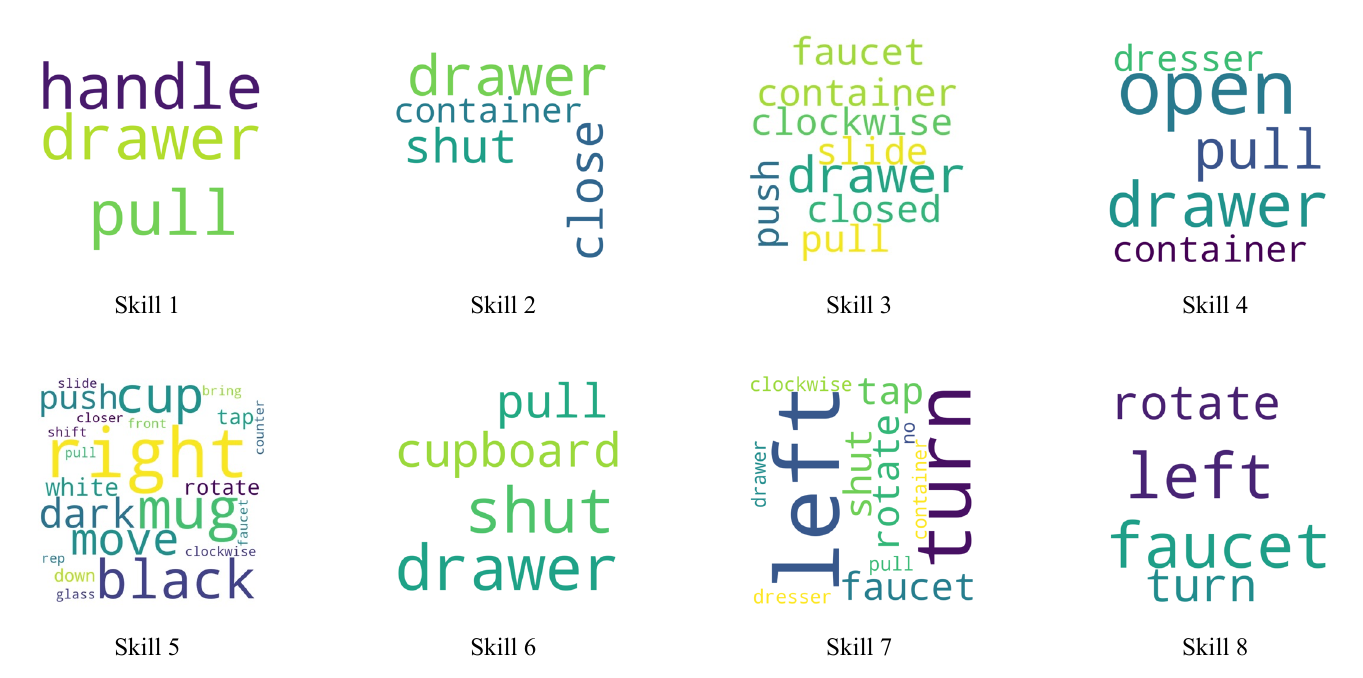}
\vspace{-10pt}
\caption{\textbf{Word cloud of learned skills in \lorl Sawyer Dataset.} We show eight of them here with the size corresponding to the word frequency in one skill.}
\vspace{-10pt}
\label{fig:wordcloud} 
\end{figure*}

\begin{figure*}[tb]
\centering 
\includegraphics[width=0.85\linewidth]{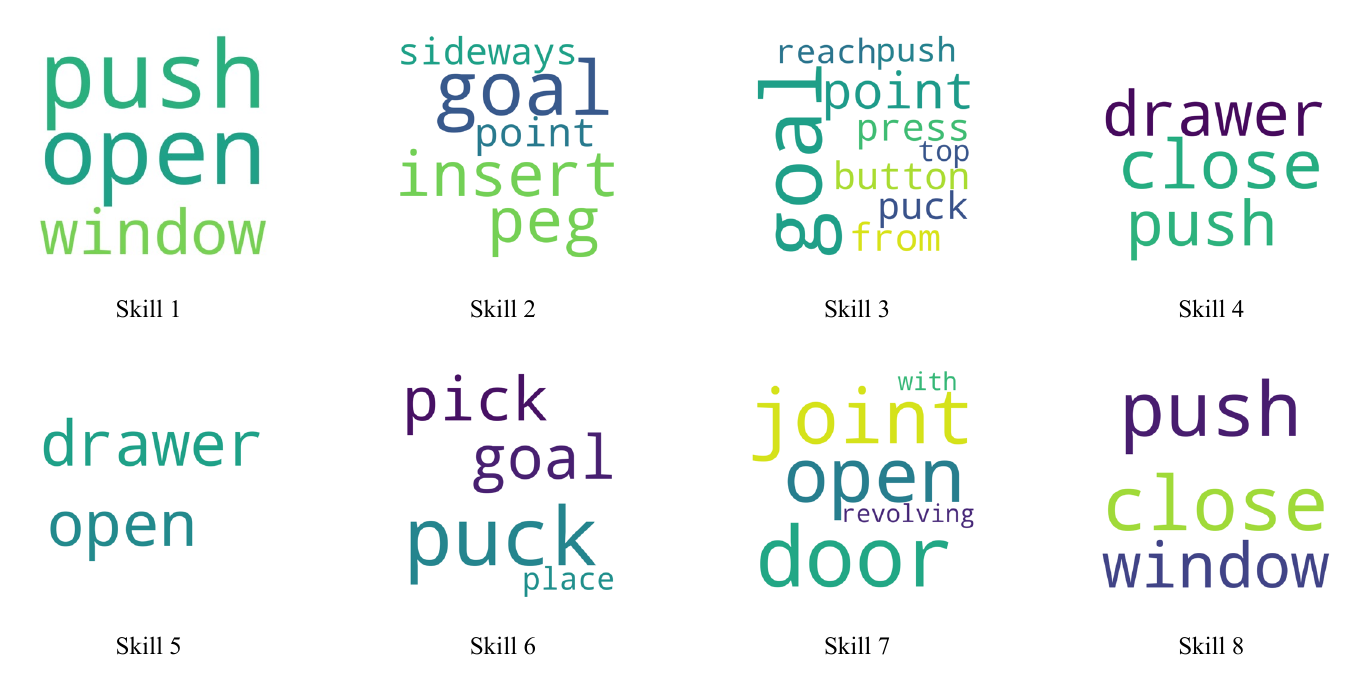}
\vspace{-10pt}
\caption{\textbf{Word cloud of learned skills in \metaworld MT10 Dataset.} We show eight of them here with the size corresponding to the word frequency in one skill.}
\label{fig:wordcloud_metaworld}
\vspace{-10pt}
\end{figure*}

We further show the word cloud of 8 learned skills of \lorl Sawyer Dataset in Figure~\ref{fig:wordcloud}. From the results, we can find that the model has successfully mastered eight key skills, each closely linked to specific tasks. These skills demonstrate strong robustness to ambiguous language instructions. For instance, skill 4 effectively abstracts the skill of ``open a drawer'' from ambiguous expressions such as ``open a container'', ``pull a dresser'', ``pull a drawer'' and random combinations of these words. Similarly, skill 6 extracts the skill of ``turn a faucet to the left''.  This analysis indicates our method's resilience to varied and poorly defined language inputs, confirming our \alias can competently interpret and act upon a wide range of linguistic instructions, even those that are ambiguous or incomplete.
% The learning and application of these two skills not only prove the advanced nature of our model in understanding and handling physical world interactions but also highlight its strong generalization ability in dealing with the ambiguity and complexity of language instructions. Furthermore, 
These findings provide new perspectives and methodological guidance for future research in similar fields, especially in handling complex tasks with ambiguous language instructions. We also provide the word cloud of learned skills from \metaworld MT10 dataset in Fig.~\ref{fig:wordcloud_metaworld}.

\section{Dataset Descriptions}
\subsection{\lorl Sawyer Dataset}
\label{append:lorl}
\begin{figure}[H]
\centering 
\includegraphics[width=\linewidth]{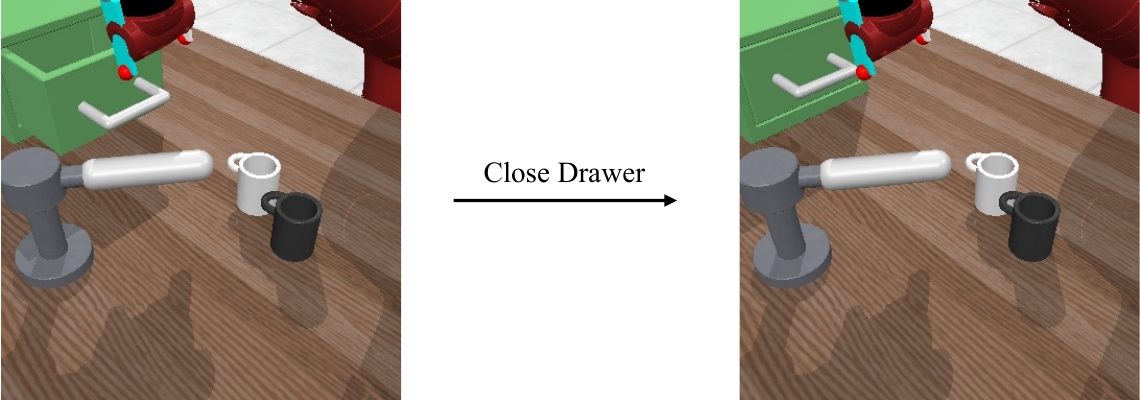}
\caption{\textbf{A sample instance of \lorl Sawyer Dataset.} The start and goal images correspond to the instruction ``close drawer''.}
\label{fig:lorel_Example} 
\end{figure}

\begin{table*}[tb]
    \centering
	\small
% 	\vskip-3pt
% 	\tabcolsep 3pt
	% \vskip-5pt
	\begin{tabular}{lp{10cm}}
 \toprule
		\textbf{Task} & \textbf{Description}\\ 
  \midrule
		Closing the Drawer                    & Involves the robot's precise manipulation of a drawer to close it, testing spatial dynamics understanding and fine motor control.                                 \\ 
\bestcell{Opening the Drawer}                    & \bestcell{Requires the robot to open a drawer, emphasizing its capability in tasks that necessitate pulling and spatial navigation.}                                             \\
Turning the Faucet Left               & Assesses the robot's precision in rotational movements for turning a faucet to the left, a nuanced everyday action.                             \\
\bestcell{Turning the Faucet Right}              & \bestcell{Tests the robot's adaptability in mirrored instructions, involving turning the faucet right, similar to the left turning task but in the opposite direction.}                                \\
Pushing the Black Mug Right           & Requires the robot to push a specific object (black mug) to the right, testing its skills in object recognition and directional movement.       \\
\bestcell{Pushing the White Mug Down}            & \bestcell{Involves pushing a different object (white mug) downward, further evaluating the robot's ability to differentiate objects and execute varied motion commands.} \\
\bottomrule
\end{tabular}
    \vspace{-5pt}
	\caption{\textbf{Overview of tasks in \lorl Sawyer Dataset.}}
 \vspace{-10pt}
 % with illustrations of each task's specific challenges and objectives
 \label{tab:lorel_dataset}
\end{table*}

Language-conditioned Offline Reward Learning dataset, abbreviated as \lorl~\cite{lorl}, contains trajectories originating from a reinforcement learning buffer which is generated by a random policy. The trajectories are sub-optimal and have language annotations through crowd-sourcing. Overall, the dataset encompasses approximately 50,000 language-annotated trajectories, each within a simulated environment featuring a Sawyer robot arm, with every demonstration extending over 20 discrete steps. A typical \lorl Sawyer environment is shown in Fig.~\ref{fig:lorel_Example}. We assess our approach using the same set of instructions as those outlined in the original paper~\cite{lorl} which are described with their objectives in Tab.~\ref{tab:lorel_dataset}. These evaluation tasks are along with various rephrases of instructions which modify either the noun (``unseen noun''), the verb (``unseen verb''), both (``unseen noun+verb''), or entail a complete rewrite of the task (``human provided''), leading to a total of 77 distinct instructions for all six tasks. This structure of tasks and rephrases enables a comprehensive assessment of the robot's ability to interpret and execute a wide range of language-based commands within the simulated environment.

\subsection{\lorl Composition Tasks}
\label{append:composition}
We follow the same settings as LISA~\cite{lisa} to create 12 new composition tasks through combining original evaluation instructions as shown in Tab.~\ref{tab:lorl-comp-instrs}.

\begin{table}[tb]
    \centering
	\small
	\begin{tabular}{c}
 \toprule
	    \textbf{Instructions}\\
     \midrule
        open drawer and move black mug right\\
        pull the handle and move black mug down\\
        move white mug right\\
        move black mug down\\
        close drawer and turn faucet right\\
        close drawer and turn faucet left\\
        turn faucet left and move white mug down\\
        turn faucet right and close drawer\\
        move white mug down and turn faucet left\\
        close the drawer, turn the faucet left and move black mug right\\
        open drawer and turn faucet counterclockwise\\
        slide the drawer closed and then shift white mug down\\
\bottomrule
    \end{tabular}
    \vspace{-5pt}
    \caption{\textbf{\lorl composition tasks}}
    \label{tab:lorl-comp-instrs}
    \vspace{-10pt}
\end{table}

Additionally, we also incorporate tasks such as ``move white mug right'' and ``move black mug down'' to explore the composition of skills related to colors (e.g., black and white) and directions (e.g., right and down). This aims to explore whether such skills can be combined to fulfill complex instructions.

\begin{table}[htb]
    \centering
	\small
% 	\vskip-3pt
% 	\tabcolsep 3pt
	% \vskip-5pt
 \resizebox{1.05\linewidth}{!}{
	\begin{tabular}{cc}
 \toprule
    \textbf{Task Identifier} & \textbf{Language Instruction}\\ 
    \midrule
    window-close & push and close a window\\ 
    window-open & push and open a window\\
    door-open & open a door with a revolving joint\\
    peg-insert-side & insert a peg sideways to the goal point\\
    drawer-open & open a drawer\\
    pick-place & pick a puck, and place the puck to the goal\\
    reach & reach the goal point\\
    button-press-topdown & press the button from the top\\
    push & push the puck to the goal point\\
    drawer-close & push and close a drawer\\
\bottomrule
\end{tabular}
}
    \vspace{-5pt}
    \caption{\textbf{Annotated instructions for \metaworld MT10 tasks.}}
    % \vspace{-15pt}
 % with illustrations of each task's specific challenges and objectives
 \label{tab:metaworld_dataset}
\end{table}

\subsection{\metaworld Dataset}
\label{append:metaworld}
The \metaworld dataset establishes a new benchmark in the field of multi-task and meta-reinforcement learning, offering 50 unique robotic manipulation tasks. These tasks range from simple to complex operations, providing researchers with a diverse testing ground. Each task is meticulously designed to ensure both challenge and common structural features that can be leveraged in multi-task and meta-learning algorithms. This design makes \metaworld an ideal choice for assessing the effectiveness and adaptability of algorithms in complex and variable task environments.

\begin{figure*}[htb]
\centering 
\includegraphics[width=0.8\linewidth]{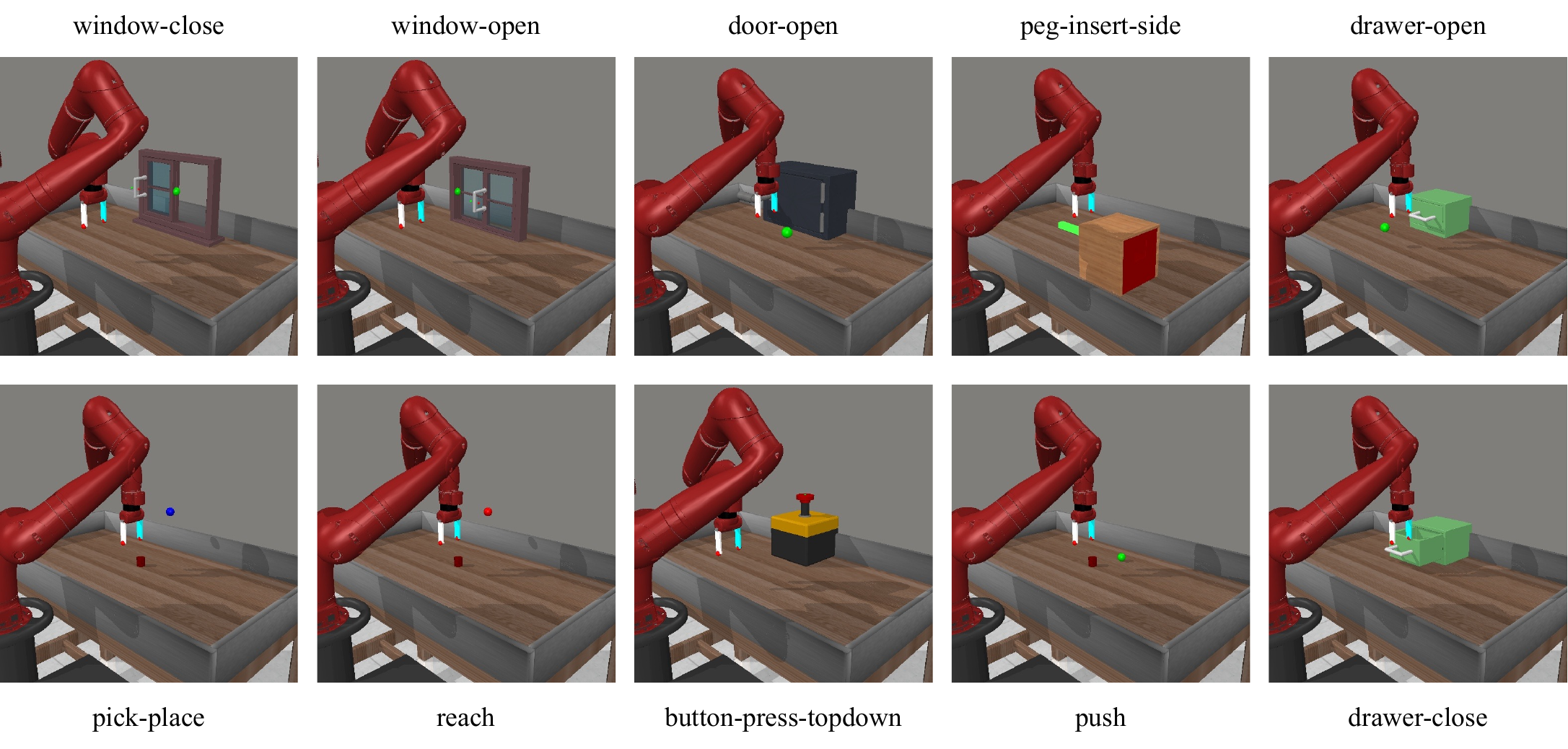}
\vspace{-5pt}
\caption{\textbf{Partially visual observations of all the 10 tasks in \metaworld MT10 Dataset.}}
\vspace{-5pt}
\label{fig:metaworld_dataset} 
\end{figure*}

Particularly, the Multi-Task 10 (MT10) subset comprises 10 carefully selected tasks, where algorithms are trained and subsequently tested on the same set of tasks. As shown in Fig.~\ref{fig:metaworld_dataset}, MT10 challenges algorithms' learning and generalization capabilities in a multi-task environment, with the aim to evaluate the consistency and efficiency of algorithms in mastering multiple tasks, as well as their adaptability and robustness in the face of diverse tasks. As there is currently no widely-recognized instruction labeling of MT10, we provide our annotations here in Tab.~\ref{tab:metaworld_dataset}.

We sample 100 trajectories for each task of MT10 and form the expert dataset of 1000 trajectories. \textcolor{cyan}{We have released our dataset with image observations on} \url{https://skilldiffuser.github.io}.

% Alongside MT10, the \metaworld dataset also features the ML10 task, a significant component designed to test meta-learning algorithms. ML10 is a more complex challenge compared to MT10, as it focuses on the concept of ``learning to learn''. In this task, algorithms are trained on a set of 10 diverse manipulation tasks, similar to MT10, but with a crucial difference. The real test in ML10 comes at the evaluation phase, where algorithms are presented with 5 new tasks that were not part of the training set. This setup is intended to assess the algorithms' ability to quickly adapt and apply learned knowledge to new, unseen tasks, a critical aspect of meta-learning. The performance on these novel tasks is indicative of the algorithms' meta-learning capabilities, making ML10 an essential part of the \metaworld benchmark for advancing research in meta reinforcement learning.

\begin{figure}[tb]
% \vspace{-2pt}
\centering 
\includegraphics[width=\linewidth]{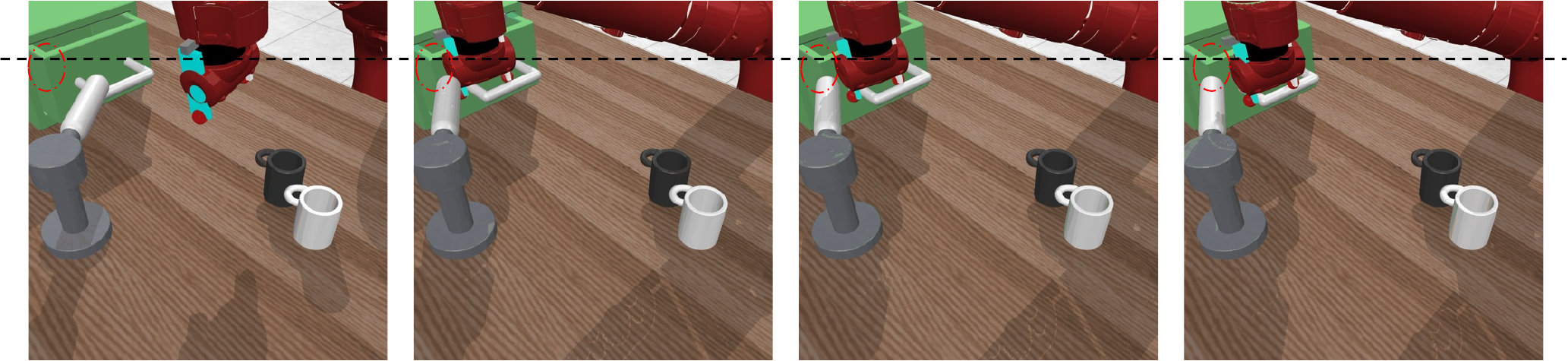}
\vspace{-15pt}
\caption{\textbf{Resulting images from applying skill 11 of Fig.~\ref{fig:heatmap_lorel}.} The black dashed line is a horizontal reference and please pay attention to the red oval region. (zoom in for best view)}
\vspace{-5pt}
\label{fig:resulting_images} 
\end{figure}

\section{More Ablations}
\subsection{Ablation Study on Skill Interpretability}
\label{append:result_image}
Resulting Images from Applying Discrete Skills
We visualize resulting images from applying skill 11 of Fig.~\ref{fig:heatmap_lorel} which has grounding of ``open, drawer,
pull, dresser, container'' (ranked from high frequency to low ones), consistent with its actual actions in Fig.~\ref{fig:resulting_images}. We can clearly observe a behavior of pulling the drawer. And we would like to clarify not all skills have clear semantic or action correspondences, while some do.

\subsection{Ablation Study on Condition Guidance Weight}
Classifier-free guidance is widely used in generative model domain for its ability to act as temperature control when setting guidance weight above 1 during inference. In all of our experiments, we set guidance weight to 1.2 by default. But we also conduct ablation study on the condition guidance weight here in Tab.~\ref{tab:guidance_weight}. From the results, we find the guidance weight slightly greater than 1 helps the planner's performance, while excessive weight hurts.

\begin{table}[htb]
\centering
\small
\resizebox{\linewidth}{!}{%
\begin{tabular}{c |c c c c c}
\toprule
\textbf{Guidance Weight} & \textbf{1.0} & \textbf{1.2} & \textbf{1.8} & \textbf{3.0} & \textbf{5.0}\\
\midrule
\textbf{Success Rate on Seen Tasks} & 39.33\% & 46.67\% & 38.86\% & 39.03\% & 33.50\%\\
\bottomrule
\end{tabular}}
  \vspace{-5pt}
  \caption{Ablation on guidance weight. (5 episodes over 3 seeds.)}
    \label{tab:guidance_weight}
    % \vspace{-10pt}
\end{table}

\begin{figure*}[htb]
\centering
\includegraphics[width=0.75\linewidth]{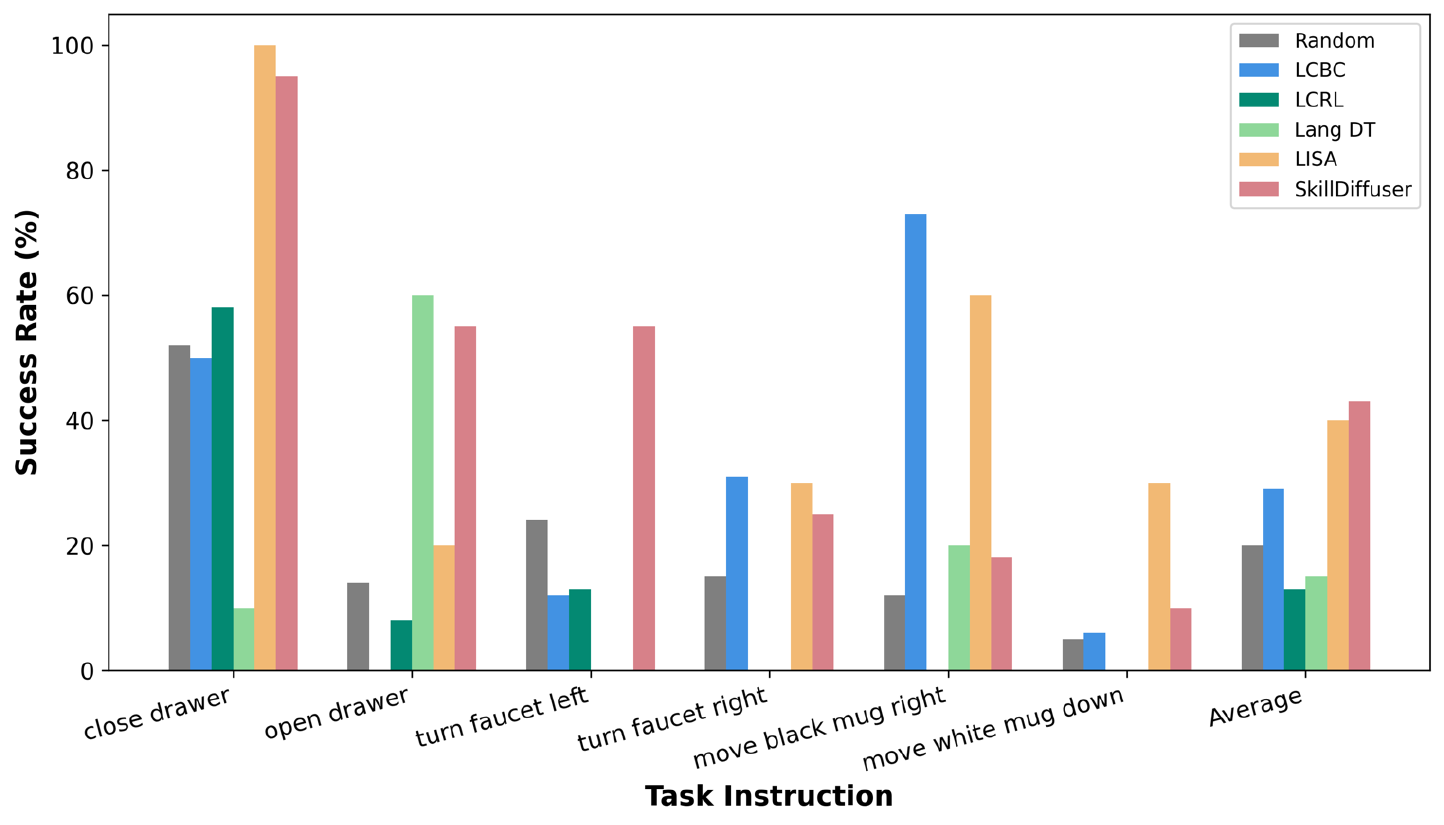}
\vspace{-8pt}
\caption{\textbf{Task-wise success rates (in \%) on \lorl Sawyer Dataset.}}
\vspace{-5pt}
\label{fig:lorel_bar} 
\end{figure*}

\begin{figure*}[htb]
\centering 
\includegraphics[width=0.75\linewidth]{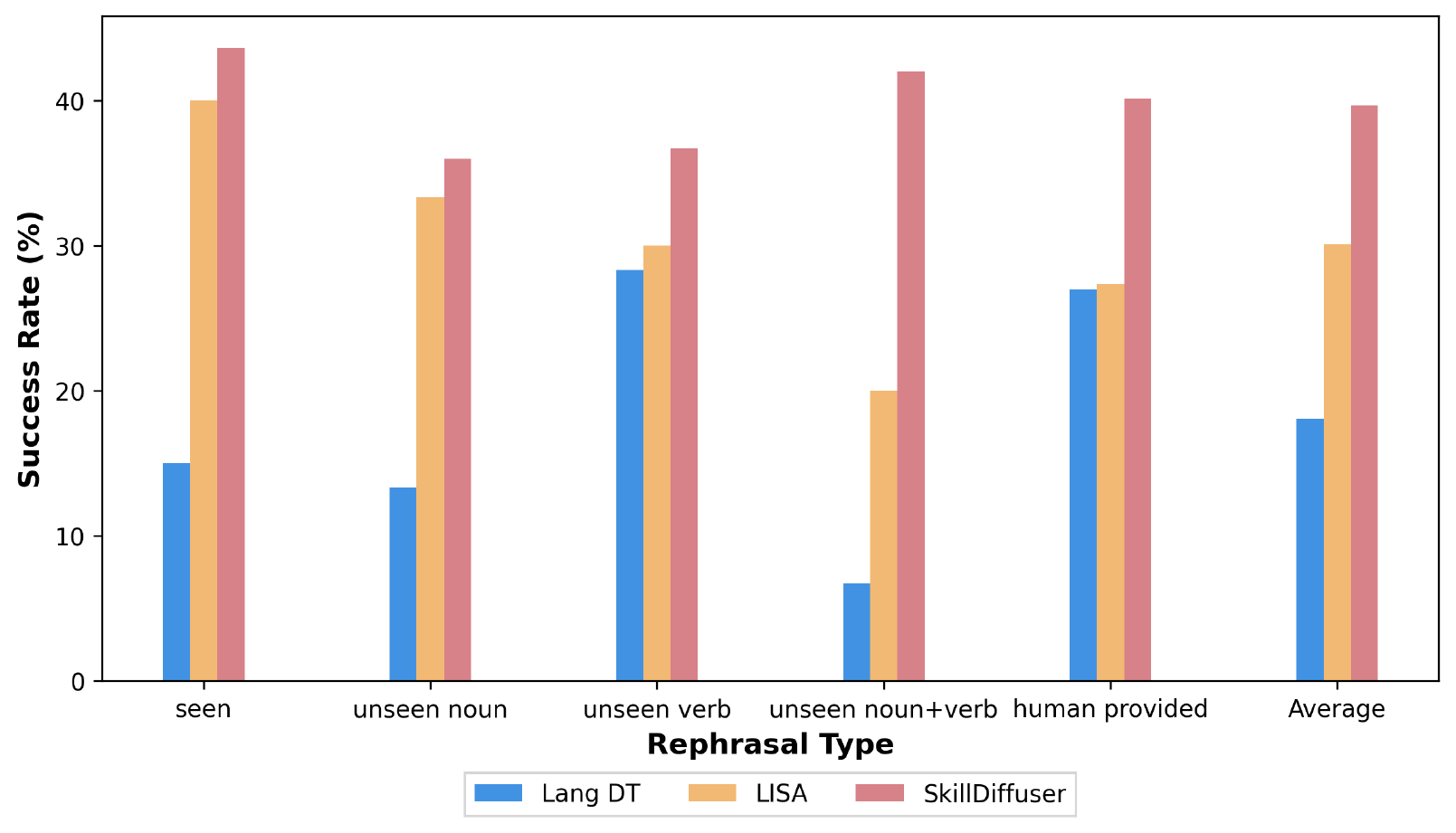}
\vspace{-8pt}
\caption{\textbf{Rephrasal-wise success rates (in \%) on \lorl
Sawyer Dataset.}}
\vspace{-5pt}
\label{fig:rephrase_bar} 
\end{figure*}

\begin{figure*}[htb]
\centering 
\includegraphics[width=0.9\linewidth]{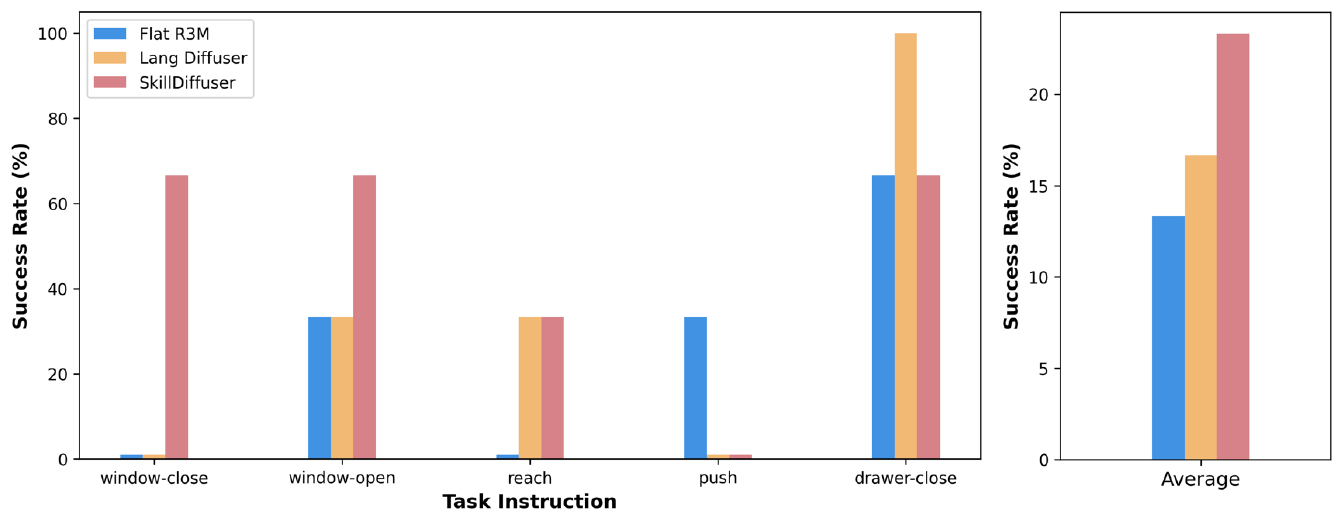}
\vspace{-10pt}
\caption{\textbf{Task-wise success rates (in \%) on \metaworld MT10 Dataset.}}
\vspace{-5pt}
\label{fig:metaworld_bar} 
\end{figure*}

\section{More Results}
\subsection{Task-wise Performance on \lorl Dataset}
We further demonstrate the performance of our method and other baselines on \lorl Sawyer dataset in Fig~\ref{fig:lorel_bar} and~\ref{fig:rephrase_bar}. As can be seen from the figures, especially from Fig.~\ref{fig:rephrase_bar}, our method's average performance on 5 rephrases is nearly 10 percentage points higher than the previous SOTA, which demonstrates its strong robustness against ambiguous language instructions. 

\subsection{Task-wise Performance on \metaworld}
We also provide the task-wise success rates on \metaworld MT10 dataset in Fig.~\ref{fig:metaworld_bar}, achieved by Flat R3M~\cite{r3m}, Language-conditioned Diffuser and \alias. The average performance is shown separately in the right figure. From our experimental outcomes, it is clear to observe that our \alias demonstrates commendable performance, particularly excelling in tasks involving mirrored instructions. \alias exhibits an average performance enhancement of over 5\% than previous language-conditioned Diffuser, which highlights the model’s advanced capability in understanding complex and ambiguous instructions compared to traditional methods. It showcases \alias's superior use of hierarchical architecture that employs interpretable skill learning for diffusion-based planners to better generate future trajectories.

\section{Implementation Details}
\label{append:train_detail}
\subsection{Hyper-parameters}
Generally, we follow the settings illustrated in~\cite{lisa} with details specified in the following Tab.~\ref{tab:hyperparameter}.

\begin{table}[H]
    \centering
	\small
	\begin{tabular}{l|cc}
 \toprule
Hyper-parameter & \lorl & \metaworld \\
\midrule
Skill Predictor Transformer Layers & 1 & 1 \\
Skill Predictor Embedding Dim & 128 & 128 \\
Skill Predictor Transformer Heads & 4 & 4 \\
Skill Set Code Dim & 16 & 16 \\
Skill Set Size & 20 & 20 \\
Dropout & 0.1 & 0.1 \\
Batch Size & 256 & 64 \\
Skill Predictor Learning Rate & 1e-6 & 1e-5 \\
Conditional Diffuser Learning Rate & 1e-3 & 5e-3 \\
Condition Guidance Weight & 1.2 & 1.2\\
Inverse Dynamics Model Learning Rate & 1e-3 & 5e-4 \\
Diffuser Loss Weight & 0.005 & 0.01 \\
Horizon & 8 & 8 \\
VQ EMA Update & 0.99 & 0.99 \\
Skill Predictor and Diffuser Optimizer & Adam & Adam \\
Inverse Dynamics Model Optimizer & Adam & Adam \\
\bottomrule
    \end{tabular}
    \vspace{-5pt}
    \caption{\textbf{Hyper-parameters of \alias.}}
    \label{tab:hyperparameter}
    \vspace{-6pt}
\end{table}

\subsection{Architecture Details}
\begin{enumerate}
    \item We use 1 layer Transformer network for the skill predictor and follow the implementation of VQ-VAE~\cite{vq-vae} to achieve VQ operation. 
    
    \item The size of skill set is set to $20$ and the planning horizon is set to $8$ for all implementations.
    
    \item A temporal U-Net~\cite{ronneberger2015u} with 6 repeated residual blocks is employed to model the noise $\epsilon_\theta$ of the diffusion process. Each block is comprised of two temporal convolutions, each followed by group norm \cite{wu2018group}, and a final Mish non-linearity \cite{misra2019mish}. Timestep and skill embeddings are generated by two separate single fully-connected layer and added to the activation output after the first temporal convolution of each block.
\end{enumerate}

\subsection{Training Details}
\begin{enumerate}
\item We train our model with one NVIDIA A100 Core Tensor GPU for about 45 hours in \lorl Sawyer dataset and about 24 hours in \metaworld MT10 dataset (1000 trajectories in total).

\item In both \lorl and \metaworld dataset, the skill predictor and diffusion model are trained with Adam optimizer~\cite{kingma2014adam} using a learning rate of $1\times10^{-3}$ for the diffusion model, $1\times10^{-6}$ for the \lorl skill predictor while $1\times10^{-5}$ for \metaworld skill predictor. We only update parameters of \metaworld skill predictor every ten iterations. The inverse dynamics model is updated with Adam optimizer as well.

\item The batch size is set to $256$ for \lorl Sawyer dataset and $64$ for \metaworld MT10 dataset.

\item The training steps of the diffusion model are $5K$ for \lorl Sawyer dataset and $8K$ for \metaworld MT10 dataset. And the training epochs of the skill predictor are $500$ for both datasets.

\item The planning horizon $T$ of diffusion model is set to $100$ and the denoising steps are set to $200$ for all tasks.

    % \item We use a planning horizon $T$ of 32 in all locomotion tasks, $128$ for block-stacking, $128$ in \texttt{Maze2D / Multi2D U-Maze}, 265 in \texttt{Maze2D / Multi2D Medium}, and 384 in \texttt{Maze2D / Multi2D Large}.
    % \item We found that we could reduce the planning horizon for many tasks, but that the guide scale would need to be lowered (\emph{e.g.}, to 0.001 for a horizon of $4$ in the \texttt{halfcheetah} tasks) to accommodate.
    % The \href{https://github.com/jannerm/diffuser/blob/34d0e93296c6d8649187e6790ee41cf0c59e3631/config/locomotion.py#L163-L178}{configuration file} in the open-source code demonstrates how to run with a modified scale and horizon.
    % \item We use $N=20$ diffusion steps for locomotion tasks and $N=100$ for block-stacking.
    % \item We use a guide scale of $\alpha=0.1$ for all tasks except \texttt{hopper-medium-expert}, in which we use a smaller scale of $0.0001$.
    % \item We used a discount factor of $0.997$ for the return prediction $\mathcal{J}_\phi$, though found that above $\gamma=0.99$ planning was fairly insensitive to changes in discount factor.
    % \item We found that control performance was not substantially affected by the choice of predicting noise $\epsilon$ versus uncorrupted data $\btau{0}$ with the diffusion model.
\end{enumerate}

\end{document}